%% file: mteb_fr.tex
\pdfoutput=1

\documentclass[11pt]{article}

\usepackage[preprint]{acl}

\usepackage{times}
\usepackage{latexsym}
\usepackage{amsmath}
\usepackage{graphicx}
\usepackage{adjustbox}
\usepackage{multirow}
\usepackage{array}
\usepackage{booktabs}
\graphicspath{ {./figures/} }

\usepackage[T1]{fontenc}

\usepackage[utf8]{inputenc}

\usepackage{microtype}

\usepackage{inconsolata}

\usepackage{listings}
\usepackage{placeins}

\title{MTEB-French: Resources for French Sentence Embedding Evaluation and Analysis}

\author{Mathieu Ciancone \\
  Wikit, France \\
  \texttt{mathieu@wikit.ai}
  \And
 Imene Kerboua \\
  Esker, France\\
  \texttt{imene.kerboua@esker.com}
  \AND
 Marion Schaeffer \\
  Wikit, France\\
  \texttt{marion@wikit.ai}
  \And
 Wissam Siblini \\
  \texttt{wissam.siblini92@gmail.com}
}

\begin{document}
\maketitle
\begin{abstract}
\input{paper_sections/abstract}
\end{abstract}

\section{Introduction}
\input{paper_sections/introduction}

\section{Related Work}
\input{paper_sections/related_work}

\section{MTEB for French}

In this section, we describe the datasets and the models that we propose for the French extension of MTEB.
We also list the research questions we want to discuss with the results.

\subsection{New Datasets}
\input{paper_sections/datasets}

\subsection{Models}
\input{paper_sections/models}

\subsection{Evaluation}
\input{paper_sections/evaluation}

\section{Results and discussion}
\input{paper_sections/results}

\section{Conclusion and perspectives}
\input{paper_sections/conclusion}

\section{Limitations}
\input{paper_sections/limitations}

\section*{Acknowledgements}
We would like to thank Wikit\footnote{\url{https://www.wikit.ai/}} and Esker\footnote{\url{https://www.esker.com/}} for providing compute and funding this research. 

\bibliography{custom}

\appendix
\input{paper_sections/appendix}

\end{document}

%% file: paper_sections/abstract.tex
Recently, numerous embedding models have been made available and widely used for various NLP tasks. 
The Massive Text Embedding Benchmark (MTEB) has primarily simplified the process of choosing a model that performs well for several tasks in English, but extensions to other languages remain challenging. 
This is why we expand MTEB to propose the first massive benchmark of sentence embeddings for French.
We gather 15 existing datasets in an easy-to-use interface and create three new French datasets for a global evaluation of 8 task categories.
We compare 51 carefully selected embedding models on a large scale, conduct comprehensive statistical tests, and analyze the correlation between model performance and many of their characteristics.
We find out that even if no model is the best on all tasks, large multilingual models pre-trained on sentence similarity perform exceptionally well. 
Our work comes with open-source code, new datasets and a public leaderboard\footnote{
French table on: \url{https://huggingface.co/spaces/mteb/leaderboard}}.

%% file: paper_sections/introduction.tex
Embeddings are dense vector representations that capture the semantics of an input. 
The first emblematic example is Word2Vec, introduced by \citet{Mikolov2013EfficientEO}. It consists of neural architectures trained to learn high-quality word representations from contextual relationships in vast amounts of text. 
Other models were proposed since then, leveraging the transformer architecture \citep{Vaswani2017AttentionIA} to produce both generic and contextualized word embeddings using self-attention.
Many models now exist with various architectures, monolingual or multilingual, pre-trained or fine-tuned \citep{naseem2021comprehensive, ding2023sentence}. 

In this work, our primary objective is to introduce a large-scale embedding benchmark for French to enable the research community and industry to select the most relevant embedding methods based on one's specific needs, such as being open-source, versatile or targeted toward a particular task, having a small embedding dimension, the ability to process long texts or their performance.
To achieve this goal, we undertake significant efforts in collecting datasets to conduct a broad comparison of models. 
We ensure that the datasets cover various tasks within a common, easy-to-use framework, and we create three new quality-checked datasets to enhance this collection. 
We select a diverse range of models, including prominent French and multilingual models deemed most efficient.
The results of our study already enable the community to make informed model selections, whether for general purposes or specific tasks. 
Additionally, our implementation is open to the community and features a public leaderboard, allowing the results to evolve with new models or datasets. With this first large-scale comparison, we perform an in-depth analysis of the results, confirming well-known findings such as the correlation between performance and model/embedding dimensions and uncovering interesting nuances.

%% file: paper_sections/related_work.tex
\paragraph{Sentence Embeddings} 
Sentence embeddings are required for many language tasks, such as Semantic Textual Similarity (STS) and knowledge retrieval. Many models have been proposed in the literature, leveraging pooling strategies \cite{Devlin2019BERTPO, muennighoff2022sgpt} or similarity fine-tuning \cite{Reimers2019SentenceBERTSE} using a contrastive framework \cite{Gao2021SimCSESC, neelakantan2022text, ni2021sentencet5, wang2022text, Zhang2023LanguageMA}, leveraging prompts \cite{wang2023improving} or a two steps training process \cite{bge-m3, lee2024nvembed}.
Few French-language models have been proposed in the literature \cite{Martin2019CamemBERTAT, le2020flaubert}. 
Most French models for sentence embeddings have been developed by the open-source community\footnote{Models on the HuggingFace hub: \href{https://huggingface.co/dangvantuan/sentence-camembert-large}{\emph{sentence-camebert}}, \href{https://huggingface.co/manu/sentence_croissant_alpha_v0.3}{\emph{sentence\_croissant\_alpha\_v0.3}}, \href{https://huggingface.co/OrdalieTech/Solon-embeddings-large-0.1}{\emph{Solon-embeddings-large-0.1}}.}, by fine-tuning models like \emph{CamemBERT}\cite{Martin2019CamemBERTAT} or \emph{CroissantLLM}\cite{faysse2024croissantllm}.

\paragraph{Benchmarks} 
Embedding models are generally compared on specific tasks, such as information retrieval, STS or reranking \cite{Thakur2021BEIR, Agirre2016Semeval, Wang2021Tsdae}. 
Other works evaluate embedding models on multiple tasks \cite{Wang2018GLUEAM, Srivastava2022BeyondTI, Conneau2018SentEval} or compare meta-embeddings \cite{Garcia2021Bench}. 
The most comprehensive benchmark to date is MTEB \cite{Muennighoff2022MTEBMT}. MTEB still has a critical limit: it mainly focuses on English. 
Some initiatives already extended this benchmark to other languages, such as Chinese \cite{xiao2024cpack} and German \cite{Wehrli2024German}. 
Our work comes with the same ambition for French. 
It relies on the MTEB structure that provides a solid basis for analysis and extends it to a new language.

%% file: paper_sections/datasets.tex
We identified 7 datasets relevant to French in the existing MTEB, which we assume are of good quality.
We complemented these with 8 external relevant datasets proposed in the literature, such as BSARD \cite{louis2022statutory} and Alloprof \cite{lef23}, which are proven to be good quality.
We created 3 new ones presented in Table \ref{tab:new_datasets} and assessed their quality with various procedures and metrics. 
In addition to all performed checks, we run multiple models on these datasets and provide results to show that they are neither trivial nor impossible to solve (see Tables \ref{tab:res_classif_pairclassif}, \ref{tab:res_reranking_retrieval}, \ref{tab:res_bitext_sts_summarization} and \ref{tab:res_clustering}).

Therefore, as of today, our French MTEB runs on 18 datasets. 
Some datasets are framed differently according to the task category they are used with. 
For example, MasakhaNEWS dataset \citep{Adelani2023MasakhaNEWSNT} is used for both Classification (\emph{MasakhaNEWSClassification}) and Clustering (\emph{MasakhaNEWSClusteringS2S} and \emph{MasakhaNEWSClusteringP2P}). 
Table \ref{tab:num_tokens} shows details of each task data used for running the benchmark. 

This section describes the 3 new datasets we introduce, quality checks performed and an analysis of the semantic similarities between datasets. 

\begin{table*}[t]
    \centering
    \resizebox{\textwidth}{!}{
    \begin{tabular}{lp{.25\textwidth}p{.25\textwidth}p{.25\textwidth}}
        \hline
        \textbf{Dataset} & \textbf{Syntec} & \textbf{HAL} & \textbf{SummEvalFr}\\
        \hline
        Samples & 100 queries \newline 90 documents & 26233 samples \newline 10 classes & 100 texts \newline 1100 human summaries \newline 1600 machine summaries \\
        Creation process & Scraping of Syntec collective bargaining agreement with articles as documents. Writing queries corresponding to articles. & Scraping of HAL articles with \emph{id}, \emph{title} and \emph{domain}. Further cleaning with deduplication, language filtering and class subsampling. & Translation from English to French with Deepl of the SummEval dataset. \\
        Annotation process & 4 annotators divided into 2 groups. Each group was given half of the articles and asked to choose an article and ask a question about it. Each annotator wrote 25 questions. & Annotations provided by authors when submitting their paper. They choose the \emph{domain} between existing academic fields. & Detailed annotation process provided in \citet{fabbri2021summeval}. \\
        Quality checks & Human verification of annotations. & Baseline models for classification and topic modeling. & Correlation between BLEU and ROUGE scores of the French and the original English datasets. LLM as-a-judge translation rating and human verification. \\
        \hline
    \end{tabular}
    }
    \caption{New datasets details with the number of samples, the creation process, the annotation process and the quality checks. All datasets are test splits.}
    \label{tab:new_datasets}
\end{table*}

\subsubsection{Syntec (Retrieval)}

    The Syntec French collective bargaining agreement\footnote{\url{https://www.syntec.fr/convention-collective/}} comprises around 90 articles. 
    Despite its topic, the language used does not feature the specificity of the legal vocabulary, making the data suitable for benchmarking general-purpose models. 
    The articles have been scraped for use as documents. 
    Four annotators were divided into two groups. 
    Each group was given half of the articles and asked to choose an article and write a question about it. 
    Each annotator wrote 25 questions. 
    Thus, a hundred questions have been manually created and paired with the articles containing the answer\footnote{\url{https://huggingface.co/datasets/lyon-nlp/mteb-fr-retrieval-syntec-s2p}}.
    Examples of the dataset are available in the appendix Figure \ref{fig:extract-syntec}.
    This dataset could also be used for text classification, clustering or topic modeling.
    Regarding quality checks, every article's integrity has been reviewed while manually creating questions. 
    We also manually checked that the questions could only be answered using the annotated article. 

\subsubsection{HAL (Clustering)}

    \emph{Hyper Articles en Ligne} (HAL) is a French open archive of scholarly documents from all academic fields. 
    Scrapping this resource, we fetched 85,000 publications in French\footnote{\url{https://huggingface.co/datasets/lyon-nlp/clustering-hal-s2s}}. 
    We extracted IDs, titles and the author's choice among domain labels.
    The last 2 are provided by authors when submitting their papers to HAL.
    Since domain annotations are provided, the dataset can be used for many tasks, such as topic modeling or text classification.
    To ensure the dataset quality is suitable for a benchmark, further data cleaning has been performed: 
    \begin{itemize}
        \item Duplicates are eliminated, retaining unique publications for each field.
        \item Irrelevant titles (due to API indexing mistakes) or titles in languages other than French have been manually removed.
        \item Samples belonging to \emph{domain} classes with less than 500 samples were removed, which leads us to keep only 10 classes. 
        \item Subsampling was performed on 2 classes containing more than 10k samples each to lower the number of samples and mitigate the unbalance of the dataset.
    \end{itemize}
    More details about this process are provided in the appendix \ref{app:hal} along with some extracts in Figure \ref{fig:extract-hal}.
    We make the dataset publicly available in both their raw and clean versions. 
    We use this dataset in a clustering setup to cluster publications by their title and use the domain as ground truth.
    To ensure the quality of this dataset, we run 3 baseline models for classification: \emph{TF-IDF + SVM}, a fine-tuned \emph{Camembert} \cite{Martin2019CamemBERTAT} and \emph{GPT-4} leveraging In-Context Learning (ICL). 
    Furthermore, we run one baseline model for topic modeling: Latent Dirichlet Allocation (LDA) \cite{blei2003latent} and report scores in the appendix \ref{app:hal}.

\subsubsection{SummEvalFr (Summarization)}
\label{sec:summeval}

    The original SummEval dataset \cite{fabbri2021summeval} consists of 100 news articles from the CNN/DailyMail dataset. 
    Each article has 11 human-written summaries and 16 machine-generated summaries annotated by 8 people with a score for coherence, consistency, fluency, and relevance. 
    We translated it from English to French using DeepL API\footnote{\url{https://www.deepl.com}}.
    Since MTEB evaluation is based on the embedding similarity between machine-generated and human-generated summaries, we propose to compute the ROUGE \cite{lin2004rouge} and BLEU \cite{papineni2002bleu} metrics between machine and human summaries for both French and English version. 
    In Table \ref{tab:rouge_bleu}, we report the average of the scores as well as their correlations between the two languages. 
    The correlation is high (above 0.7), showing that the word and n-gram overlap between human and machine summaries is highly preserved in the French version.
    One may argue that computing the metric on fully translated texts (human and machine summaries are both translated from English) may introduce biases and not assess the quality of the translations. 
    For this purpose, we ensure the French human summaries are correctly translated from English. 
    We use an LLM as-a-judge \cite{zheng2023judging} where given the original human summary in English and its translation in French, the model rates the quality of the translation from $0$ to $10$, with $0$ being of very bad quality and $10$ being excellent. 
    The prompt is available in Figure \ref{fig:prompt-llm-judge}.
    Additionally, we manually check random translations with ratings between 9 and 10 to ensure the rating is relevant. 
    We do the same for all translations with a score less than $9$ and correct them\footnote{SummEvalFr available at: \url{https://huggingface.co/datasets/lyon-nlp/summarization-summeval-fr-p2p}} (see the rating distribution in Table \ref{tab:ratings}). 

\begin{table}[htbp]
    \centering
    \resizebox{\columnwidth}{!}{
    \begin{tabular}{lcccc}
        \hline
        \textbf{Dataset} & \textbf{BLEU} & \textbf{ROUGE-1} & \textbf{ROUGE-2} & \textbf{ROUGE-L}  \\
        \hline
        SummEval & 0.205 & 0.292 & 0.099 & 0.193 \\
        SummEvalFr & 0.276 & 0.302  & 0.117 & 0.194 \\
        Correlation En-Fr & 0.70 & 0.85 & 0.80 & 0.84 \\
        \hline
    \end{tabular}
    }
    \caption{Average ROUGE and BLUE scores computed between machine summaries and human summaries for the original English SummEval and its translation to French. The correlations of the individual scores between English and French are also reported.}
    \label{tab:rouge_bleu}
\end{table} 

\subsubsection{Data for the Reranking task}

The reranking task, as evaluated in MTEB, requires datasets composed of a set of queries, each associated with relevant and irrelevant documents. 
Despite our efforts, we found no French dataset that natively exhibits such a structure. 
Thus, to evaluate this task, we built data for the reranking task based on the \textit{Syntec} and \textit{Alloprof} \cite{lef23} datasets. 
These already feature queries and labeled relevant documents. 
Irrelevant ones were added using the following process:  
\begin{itemize}
    \item To avoid bias, we use the BM25 algorithm \cite{Robertson1976RelevanceWO} (which is a deterministic method) to rank documents in terms of relevance regarding each query. 
    \item The top 10 documents that are not labeled as relevant constitute the negative samples.
\end{itemize}
We recognize that this process leads to a high correlation between the retrieval and reranking tasks. 
We still think it is essential to make the latter available, with an open door to future improvement\footnote{SyntecReranking available at: \url{https://huggingface.co/datasets/lyon-nlp/mteb-fr-reranking-syntec-s2p} and AlloprofReranking available at: \url{https://huggingface.co/datasets/lyon-nlp/mteb-fr-reranking-alloprof-s2p}}. 

\subsubsection{Similarity analysis}\label{sec:dataset_sim}

We investigate the proximity between the datasets' topics to give insights about the benchmark contents. 
The methodology introduced by \citet{Muennighoff2022MTEBMT}, i.e. computing an average embedding of samples from each dataset, is used to build a dataset-similarity matrix (displayed in appendix Figure \ref{fig:cosine_similarity}).
The distances between averaged embedding vectors of each dataset (which range from $0.89$ to $1$ in Figure \ref{fig:cosine_similarity}) remain hard to interpret into a dataset semantic proximity. Thus, we complement this by observing the dataset's clouds of embedding in a 2D plane using PCA in Figure \ref{fig:PCA_similarity}.

Figures \ref{fig:PCA_similarity} and \ref{fig:cosine_similarity} seem to correlate, showing high similarity between two datasets when the same underlying data is used in different tasks. 
Dataset topics are pretty close, with some exceptions, such as the Syntec dataset.
As more datasets are added to the benchmark, this analysis will help select new data that do not produce redundant results.
It may also help to understand the link between the results and the datasets' topics.

%% file: paper_sections/models.tex
\label{sec:models}
For comparison on our benchmark, we selected various models to fulfil three objectives.
\begin{itemize}
    \item \textbf{Quantity:} The aim was to compare a substantial number of models (51 in total) to provide comprehensive results, facilitating the community in selecting effective French models. 
    \item \textbf{Relevance:} It was imperative to include top performers from the MTEB benchmark \citep{Muennighoff2022MTEBMT}. We mainly selected multilingual models and some English models to asses their language-transferring abilities. Additionally, we integrated natively French transformer-based models such as \emph{CamemBERT} \citep{Martin2019CamemBERTAT}, \emph{FlauBERT} \citep{le2020flaubert} and even the very recent \emph{CroissantLLM} \cite{faysse2024croissantllm}.
    \item \textbf{Variety:} Diverse model types were included to offer an insightful analysis across various model characteristics (dimension, training strategy, etc.).
\end{itemize} 

In line with the third objective, we explicit below the studied characteristics of embedding models that will be discussed with the results.
\begin{itemize}
    \item \textit{\textbf{Embedding dimension:}} This critical element influences the expressiveness of the representation and, in practical applications, the underlying storage and compute costs. We selected models with embedding dimensions ranging from 384 to 4096.
    \item \textit{\textbf{Sequence length:}} Being the number of tokens that a model can consider as input, the sequence length is important as it impacts the unit that can be encoded (sentence, paragraph, document). However, encoding overly long sequences requires efficiently storing the relevant information into a single vector. Among the selected methods, this criterion varies from 128 tokens to 32768.
    \item \textit{\textbf{Model parameters:}} Often correlated with the two first characteristics, parameter count is important for practical applications as it affects usability on resource-efficient machines. The selected models have a number of parameters ranging from 20 million ($\sim$100Mb in float32) to 7 billion ($\sim$28Gb).
    \item \textit{\textbf{Language:}} This is a major feature of language models. Some are monolingual, and others are multilingual. Language is usually acquired during pre-training, but sometimes, models familiarize themselves with new languages at tuning. For the benchmark, we selected French models, as well as bilingual or multilingual models. We also included a few ones that claimed to be English (e.g. \emph{all-MiniLM-L12-v2}\footnote{\url{https://huggingface.co/sentence-transformers/all-MiniLM-L12-v2}}).
    \item \textit{\textbf{Model types:}} There are several strategies to generate text embeddings such as aggregating (e.g. with average pooling) token-level embeddings from raw pre-trained models, or adding an extra contrastive learning step on a sentence similarity task with, optionally, additional transformation layers. We included models of all types in our benchmark, summarizing the model type information under two relevant criteria: finetuned vs pretrained, and trained for sentence similarity or not.
\end{itemize}

The selected models are visible in Figure \ref{fig:crit_diff_diagram}, and all of their characteristics are summarized in appendix Table \ref{tab:model_characteristics}. Overall, the selection includes the best models from the sentence transformers framework \citep{Reimers2019SentenceBERTSE}, the most popular French NLP models \cite{le2020flaubert, Martin2019CamemBERTAT}, their variants optimized for semantic similarity \cite{Reimers2019SentenceBERTSE}, numerous multilingual models performing at the top on MTEB (e.g \emph{E5} and \emph{T5}), \emph{Bloom} variants \cite{ Zhang2023LanguageMA}, models based on very recent powerful LLMs \cite{wang2023improving, faysse2024croissantllm} and finally the proprietary models of OpenAI, Cohere and Voyage. Certain models were selected in multiple sizes to isolate the dimensionality effect effectively. We provide information on the models' licenses as reported in the Hugging Face hub\footnote{\url{https://huggingface.co/models}}. However, we encourage readers to conduct further research before utilizing a model.

%% file: paper_sections/evaluation.tex
For the sake of homogeneity, models are evaluated using the same metrics per task as in MTEB \cite{Muennighoff2022MTEBMT}: Classification (Accuracy), Bitext mining (F1 score), Pair classification (AP), Clustering (V measure), Reranking (MAP), Retrieval (NDCG@10), Summarization and STS (Spearman correlation based on cosine similarity).
BitextMining tasks are excluded from the average performance scores and therefore the figures, as this task evaluates 2 languages instead of one, and this benchmark focuses only on one language (French). We present the results for both \emph{DiaBlaBitextMining} and \emph{FloresBitextMining} in Table \ref{tab:res_bitext_sts_summarization}.

Using the overall benchmark results, our goal will be to answer the following research questions:
\\
\textbf{Q1: }Is a model outstanding on all tasks? 
\\
As we are trying to find out whether one embedding model is statistically better than the others for French, the objective will also be to analyze the performance of the models by tasks to facilitate model choice for specific applications. 
\\
\textbf{Q2: }Are there any links between the model characteristics and performance? 
\\
In section \ref{sec:models}, we undertook the substantial task of gathering the characteristics of all evaluated models. The goal here will be to analyze their impact on performance and draw conclusions about, for example, the relationship between embedding dimension and model ranking on the benchmark.
\\
\textbf{Q3: }Do monolingual models have multilingual capabilities?
\\
We interrogate the ability of a model trained exclusively in one language to perform well in another language.
\\
\textbf{Q4: }Are there any correlations between datasets with respect to model ranking? 
\\
To go further than the correlation analysis among datasets regarding their topics (see section \ref{sec:dataset_sim}), subsequent analysis will be conducted regarding how they rank models. Additionally, complementary insights will be derived from examining correlations of models relative to their strengths and weaknesses across different datasets.

%% file: paper_sections/results.tex
In this section, we present the results through the prism of our research questions.


\subsection*{Q1: Is there a model that outstands on all tasks?}

\begin{figure*}[ht]
    \centering
    \includegraphics[width=1\textwidth]{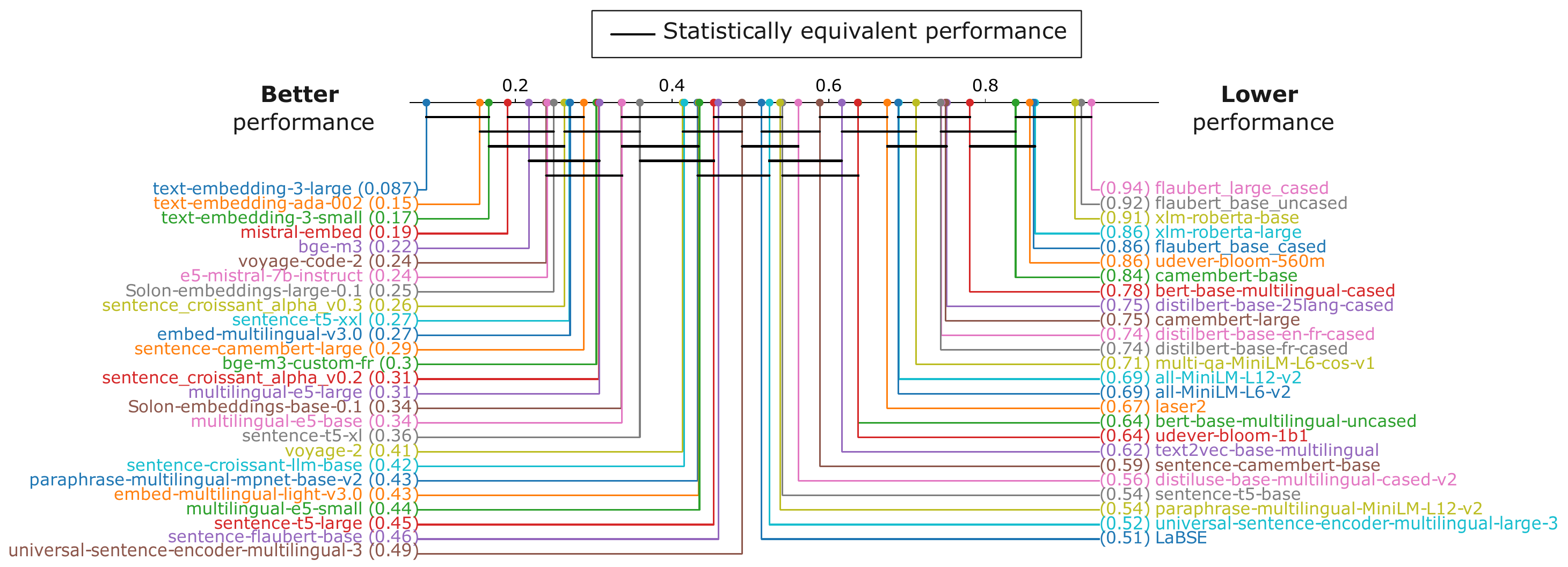}
    \caption{Critical difference diagram representing the significant rank gaps between models. The axis represents the normalized average rank of the models (lower is better). The black bars indicate that the difference in models' rank is not statistically significant, i.e. lower than the critical difference.}
    \label{fig:crit_diff_diagram}
\end{figure*}

Models performances for each task are presented in appendix Tables \ref{tab:perf_per_task_type}, \ref{tab:res_classif_pairclassif}, \ref{tab:res_reranking_retrieval}, \ref{tab:res_bitext_sts_summarization} and \ref{tab:res_clustering}. 
Figure \ref{fig:crit_diff_diagram} shows the critical difference diagram of average score ranks. 

As in MTEB \cite{Muennighoff2022MTEBMT}, no model claims state-of-the-art in all tasks even if the \emph{text-embedding-3-large} model is in first place on average on all tasks (see Table \ref{tab:perf_per_task_type}). 
It ranks first for the classification and reranking tasks.
For the clustering task, \emph{text-embedding-ada-002} is the best model.
The models \emph{voyage-code-2}, \emph{text-embedding-3-small} and \emph{mistral-embed} share the top positions in the retrieval task ranking.
For the pair classification task, \emph{laser2} is ahead of its competitors.
Finally, \emph{sentence-camembert-large} leads on the STS task and \emph{multilingual-e5-small} has the best results for summarization.

Figure \ref{fig:crit_diff_diagram} shows a global model comparison across all datasets.
The models are arranged horizontally according to their performance, with the best models on the left.
The black bars represent the statistical equivalence between the models' performances. 
The statistically equivalent top performers for this benchmark are OpenAI's models \emph{text-embedding-3-large}, \emph{text-embedding-3-small} and \emph{text-embedding-ada-002}.
Interestingly, many models do not show a significant performance gap between their base and large flavours.
Some French models stand out among the multilingual models, such as \emph{Solon-embeddings-large-0.1}, \emph{sentence\_croissant\_alpha\_v0.3} and \emph{sentence-camembert-large}.

\subsection*{Q2: Are there any links between model characteristics and performance?} 
\label{sec:results-Q2}

The Spearman correlations between the average rank of the models and their characteristics are the following: 
\begin{itemize}
    \item \emph{Tuned for sentence similarity}: 0.727
    \item \emph{Finetuned vs pretrained}: 0.544
    \item \emph{Model number of parameters}: 0.49
    \item \emph{Embedding dimension}: 0.452 
    \item \emph{Closed source}: 0.449
    \item \emph{Max sequence length}: 0.336  
    \item \emph{Multilingual}: 0.103 
    \item \emph{English}: 0.025
    \item \emph{English but tuned on other languages}: -0.025
    \item \emph{French}: -0.134
    \item \emph{Bilingual}: -0.135
\end{itemize} 
Additionally, all cross-correlations between characteristics are reported in appendix Figure \ref{fig:cross_correlation_characs}. 

As expected, the score most strongly correlates with whether the evaluated models were trained on a sentence similarity task. Of course, this criterion is connected to the more general \emph{Finetuned} one. The only top-performing models solely pre-trained are from the \emph{E5} family, where the pre-training is, in fact, contrastive and optimized for similarity. Conversely, models pre-trained on token-level tasks and generating embeddings via pooling appear less well-suited for the benchmark tasks.

Furthermore, we observe a performance correlation with the embedding dimension and the model's number of parameters, which are often correlated themselves. This appears very clearly on the relative ranking of \emph{E5} and \emph{T5} models (see Figure \ref{fig:crit_diff_diagram}). However, some small models perform very well on the benchmark, such as the standard version of the multilingual universal sentence encoder or \emph{Solon-embeddings-base-1.0}. Notably, the maximum sequence length, while an important criterion for generative tasks with LLMs, is less correlated with performance than the other dimensions. This can be explained by many datasets containing relatively small texts (see appendix Table \ref{tab:num_tokens} showing that 14 datasets have less than 50 tokens).

Regarding language, it is surprising that good performance is not particularly correlated with French models in particular. In reality, the other aspects of the models, such as being fine-tuned for similarity, prevail. Nevertheless, we can highlight the excellent performance of a few French models such as \emph{sentence-camembert} and \emph{sentence-croissant} and \emph{Solon-embeddings}. 

Lastly, we emphasize that closed-source models perform well on this benchmark (\emph{text-embeddings}, \emph{mistral-embed} and \emph{voyage}), but we lack information about their characteristics. As more open-source well-performing models get added in the future, we could expect this correlation to decrease. Note that the correlation between sequence length and performance could be dragged by closed-source models that have generally larger sequence lengths.

\subsection*{Q3: Do monolingual models have multilingual capabilities?}

\begin{figure}[!ht]
    \centering
    \includegraphics[width=\columnwidth]{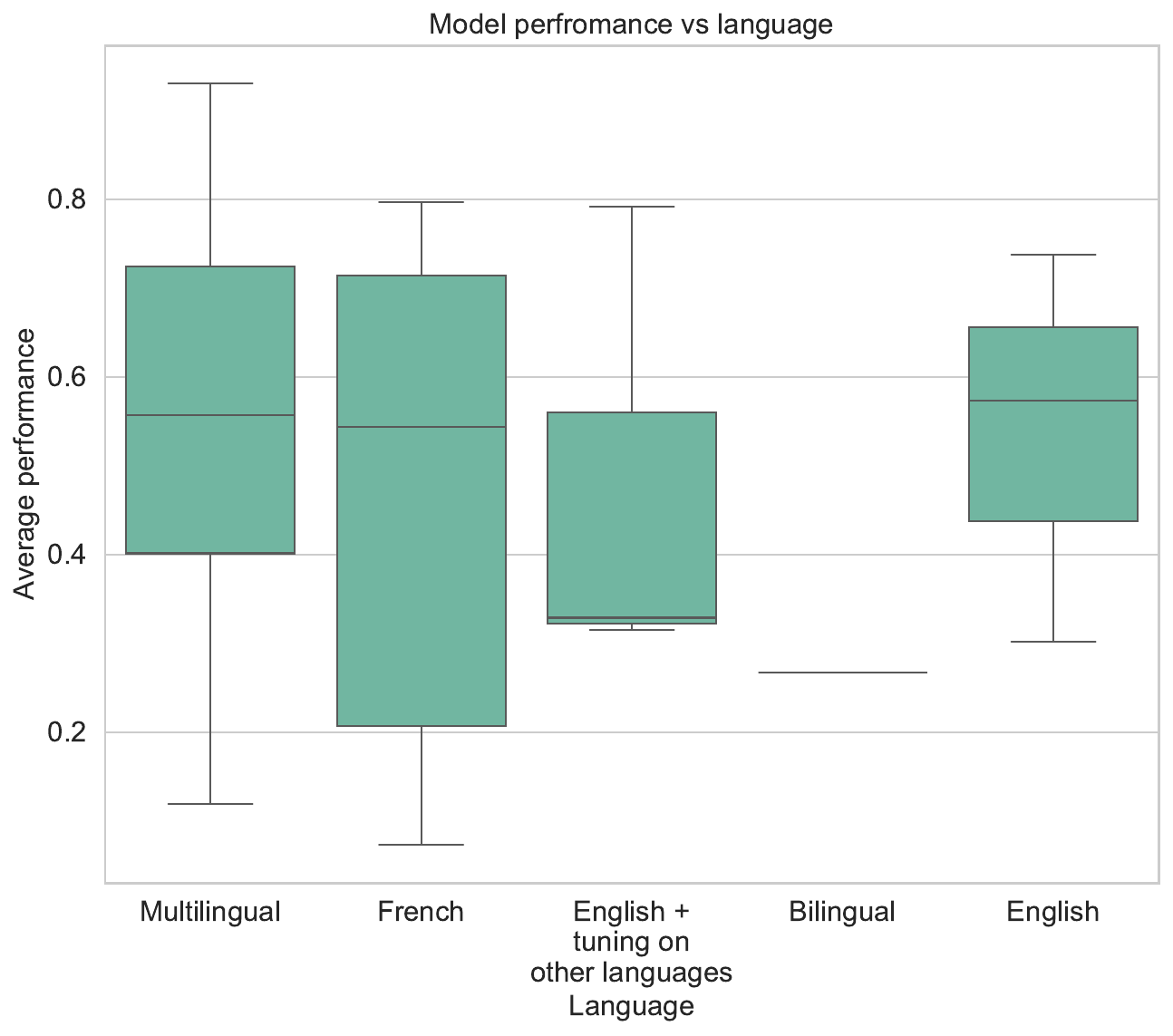}
    \caption{Model performance depending on the language of the data they have been trained on.}
    \label{fig:perf_per_lang}
\end{figure}

We also studied the capabilities of models on the French language when the language of the training data varies.
It is surprising to note the absence of a clear correlation between the language the model is trained on and its performance on French, as shown by the large standard deviation in Figure \ref{fig:perf_per_lang}.\\
Furthermore, monolingual models trained exclusively on English such as \emph{voyage-code-2} show very good results on French datasets compared to models trained exclusively on French such as \emph{flaubert} derivatives and \emph{distilbert-base-fr-cased} (see Table \ref{tab:appendix_avg_results}).
\\
This is explained by the fact that a large part of the selected French models generate embeddings using a pooling strategy. Only a few are sentence transformer models, for which the pooled representation is part of the model and trained with it, leading to higher-quality embeddings. This is endorsed by the excellent results of \emph{sentence-camembert-large}, a sentence transformer model trained on French corpus and confirms the recent findings in terms of model architecture \cite{Gao2021SimCSESC}.
\\
Finally, it should be noted that a significant portion of the French data used to train the selected French models actually comes from English datasets that have been machine translated \citep{translated_sts}. Despite the tremendous progress of machine translation, it is well known that the generated data may be unrepresentative of the language used by native speakers and cause a reduced final performance \citep{impact_automatic_translation}.
\\

\subsection*{Q4: Are there any correlations between datasets with respect to model ranking?} 

The datasets correlation w.r.t model ranking are presented in appendix Figure \ref{fig:correlation_dataset_results}. Except for two datasets (\emph{MasakhaNEWSClusteringP2P}, \emph{SummEvalFr}), the correlations, on average, are high. There is still enough diversity to make each dataset interesting for the French MTEB benchmark. Two groups (\emph{SyntecReranking}/ \emph{SyntecRetrieval}, \emph{MassiveScenarioClassification}/ \emph{MTOPDomainClassification}/ \emph{MassiveIntentClassification}) exhibit notably high correlations ($\sim$0.97).
It is interesting to point out some sub-diagonal correlation blocks. The datasets being arranged by task indicate that models behave slightly more similarly within the same task than between two different tasks. This underscores the importance of having multiple tasks in the benchmark to select general-purpose models. For readers interested in specific tasks, it is more relevant to examine task-specific rankings rather than the overall one.
The complementary results of model correlations w.r.t to strengths and weaknesses on datasets are displayed in appendix Figure \ref{fig:correlation_model_results}. Strong correlations in behavior emerge among the variants of the same models (e.g. DistilBERT, sentence-croissant, sentence-t5, e5, etc.). Correlations are also generally observed among numerous models trained using the sentence transformers framework \cite{Reimers2019SentenceBERTSE}, as well as proprietary models, e.g. from Cohere and OpenAI. Conversely, these models fine-tuned for sentence similarity, show minimal correlation with pre-trained models for which token-embedding pooling techniques are employed.

%% file: paper_sections/conclusion.tex
In this work, we introduce a large-scale embedding benchmark for French to enable the research community and industry to select the most relevant embedding methods based on their specific needs.
We undertake significant efforts in collecting 15 datasets and create 3 new quality-checked ones to enhance this collection.
The whole French benchmark runs on 26 tasks.
We select a diverse range of 51 models, including prominent French and multilingual models deemed most efficient to conduct a broad comparison.
Our implementation is open to the community and features a public leaderboard, allowing the results to evolve with new models or datasets. 
After an in-depth analysis of the results, OpenAI models perform significantly better than the other models. 
However, other models should be considered for their performance on specific tasks, being open source or having a small embedding dimension. 


This work opens several doors for future improvements. 
By examining dataset diversity in terms of topics and model ranking, we observe that the benchmark would benefit from additional datasets that introduce higher diversity. 
Beyond classification, many tasks focus on semantic similarity, explaining the strong performance of models trained for similarity. 
Exploring novel tasks in the generative spectrum or evaluating token embeddings (contextualized or not) on tasks like Named Entity Recognition could be an interesting path for future exploration. 
There are also opportunities for improvements on the model side. 
With numerous existing models that could be added to the leaderboard and many new proposals awaiting. 
For instance, we can already see the promising capabilities of early variants of recent models \cite{faysse2024croissantllm} and expect that future proposals will come to compete strongly with closed-source models.
Ultimately, we hope to see the emergence of other language-specific MTEB variants (e.g. for high-resource languages like Spanish and German), enabling a more comprehensive evaluation of multilingual model performance.

%% file: paper_sections/limitations.tex
\paragraph{Native French resources unavailability}
The availability of resources natively in French is an obvious limitation of our work. 
Regarding models, there are far fewer options than with more widespread languages such as English. 
Indeed, most of the existing French embedding models we found are trained using either older architectures or methods, unlike most recent multilingual models such as \emph{NV-Embed-v1} \citep{lee2024nvembed} or \emph{e5-mistral-7b-instruct} \citep{wang2023improving}. 
Comparing models by family would be beneficial, particularly for evaluating French models against multilingual models on the same architecture using the same training technique. 
Resource limitations also apply to datasets. 
For example, the summarization task dataset is translated, which can be less relevant than a natively French dataset. 
We have also built datasets for reranking tasks using existing ones from retrieval task because we could not find any in French. 
This construction process introduces a bias as the model performance on both tasks may be correlated (see Figure \ref{fig:correlation_dataset_results}). 
We preferred to propose datasets even if they could introduce biases rather than not address the task in the benchmark. Note that each task type can be considered individually.
We hope additional resources will be developed in the French-speaking community to enrich our comparison. 

\paragraph{Benchmark validity over time}
As with all benchmarks, their reliability over time can be discussed as the field evolves fast.
The models selected for the analysis conducted in this paper are those available at this time, new outperforming models will be created and shall be evaluated.
Our work extends MTEB and thus simplifies the addition of new datasets for evaluation and allows running new models.
With this effort, we hope this will simplify the evaluation of new models proposed by the community to keep our work up to date.

\paragraph{Data contamination issues} 
Bias may exist for models that use the training sets of the provided evaluation datasets for their training. 
It considerably improves their performance on the benchmark, favouring them over other models. 
This is particularly worrying for models that do not communicate about the datasets used during training, such as proprietary models. 
Generally speaking, it would be interesting to calculate the similarity between the datasets used to train the models and those used to test them to check that they are far enough apart to draw general conclusions.

\paragraph{Focus on sentence embeddings}
Finally, like the original version of MTEB, the comparison focuses mainly on sentence embeddings. Other tasks could be added to cover word embeddings and, therefore, more NLP tasks. 

%% file: paper_sections/appendix.tex
\newpage
\label{sec:appendix}

\section{Supplementary materials for datasets}

\subsection{All datasets}

Table \ref{tab:num_tokens} displays the size of each dataset along with the average number of tokens per sample and their references. The dataset's content was tokenized using \emph{cl100k\_base} encoding. For Retrieval, the two numbers refer to the queries and the documents. For Reranking, the three numbers refer to the queries, the pairs of queries with relevant documents and the pairs of queries with irrelevant ones, respectively. The pairs of queries and documents are obtained from the 90 documents extracted. For \emph{SummEvalFr}, the three numbers refer to the texts, human and machine summaries, respectively.

\input{tables/num_tokens_table}

\begin{figure*}[htbp]
    \centering
    \includegraphics[width=0.95\linewidth]{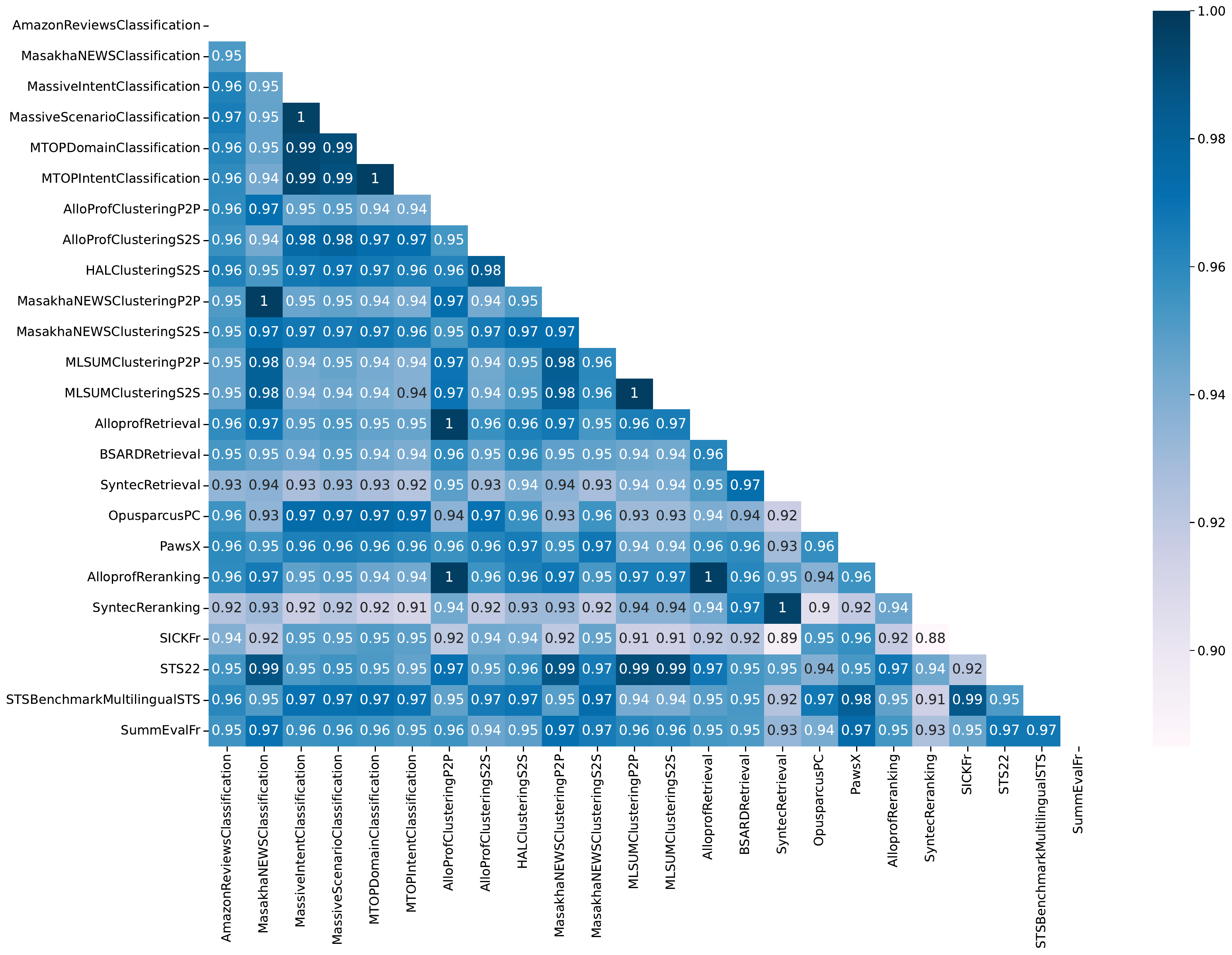}
    \caption{Cosine similarity between tasks' data. Ninety random samples per task's data are embedded using the \emph{multilingual-e5-small} model. The embeddings of each task's data sample are averaged. The similarity between each dataset is then calculated using cosine similarity as in \cite{Muennighoff2022MTEBMT}.}
    \label{fig:cosine_similarity}
\end{figure*}

Figure \ref{fig:cosine_similarity} represents the semantic similarity between each dataset. The methodology was as follows: 90 random samples per dataset are embedded using the \emph{multilingual-e5-large} model. The embeddings of each dataset's samples are averaged. The similarity between each dataset is then calculated using cosine similarity as in \cite{Muennighoff2022MTEBMT}.

\begin{figure*}[htbp]
    \centering
    \includegraphics[width=0.9\textwidth]{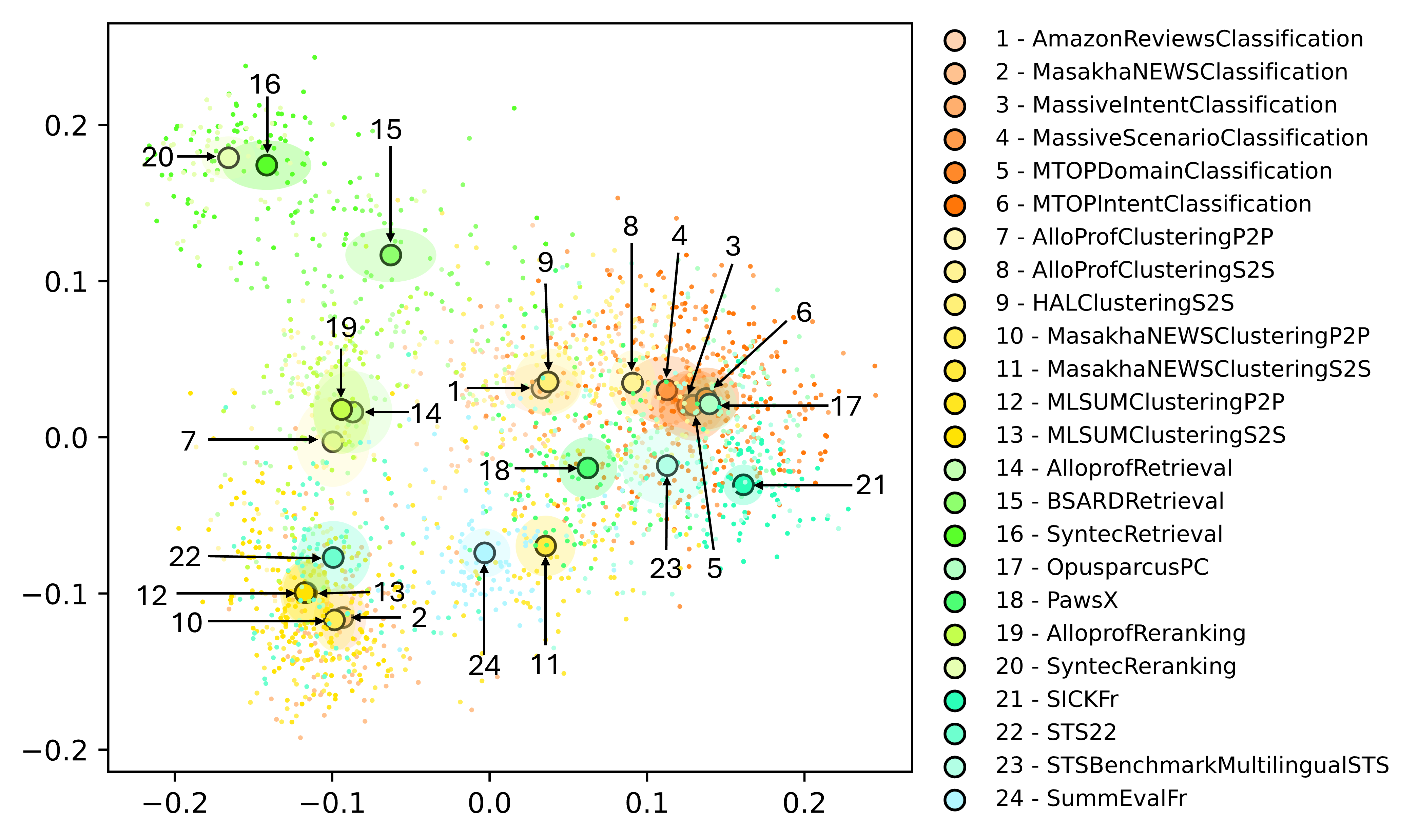}
    \caption{2D projection of tasks' data. 90 random samples per task's data are embedded using \emph{multlingual-e5-small} model \cite{wang2022text}. The embeddings are reduced to 2 dimensions using PCA. The centroid of each task's data is represented, along with the ellipse showing the standard deviation along each axis.}
    \label{fig:PCA_similarity}
\end{figure*}

We complement this analysis by observing the dataset’s clouds of embedding in a 2D plane using PCA in Figure \ref{fig:PCA_similarity}.

\subsection{Created datasets}

\paragraph{Syntec}

Figure \ref{fig:extract-syntec} shows an extract from the Syntec dataset with a document and a query relative to this document.

\begin{figure}[htbp]
\centering
\textbf{Document}

\begin{tabular}{|c|p{0.7\linewidth}|}
\hline
 id & article-14 \\ 
 \hline
 url & https://www.syntec.fr/convention-collective/resiliation-du-contrat-de-travail/\#article-14 \\  
 \hline
 title & Article 14 : Préavis pendant la période d’essai \\
 \hline
 section & Résiliation du contrat de travail \\
 \hline
 content &  Modification Avenant n° 7 du 5/07/1991 Au cours de cette période, les deux parties peuvent se séparer avec un préavis d’une journée de travail pendant le premier mois. Après le premier mois, le temps de préavis réciproque sera d’une semaine par mois complet passé dans l’entreprise. Après le premier mois, le temps de préavis réciproque sera d’une semaine par mois passé dans l’entreprise. Le préavis donne droit au salarié de s’absenter pour la recherche d’un emploi dans les conditions fixées à l’article 16. Le salarié sera payé au prorata du temps passé pendant la période d’essai. \\
\hline
\end{tabular}
\textbf{Query}
\begin{tabular}{|c|p{0.7\linewidth}|}
\hline
 article & article-14 \\ 
 \hline
 question & Quel est le préavis en période d'essai ? \\  
\hline
\end{tabular}
\caption{Extracts of Syntec dataset.}
\label{fig:extract-syntec}
\end{figure}

\paragraph{HAL}
\label{app:hal}

\begin{figure}[htbp]
    \centering
    \includegraphics[width=\columnwidth]{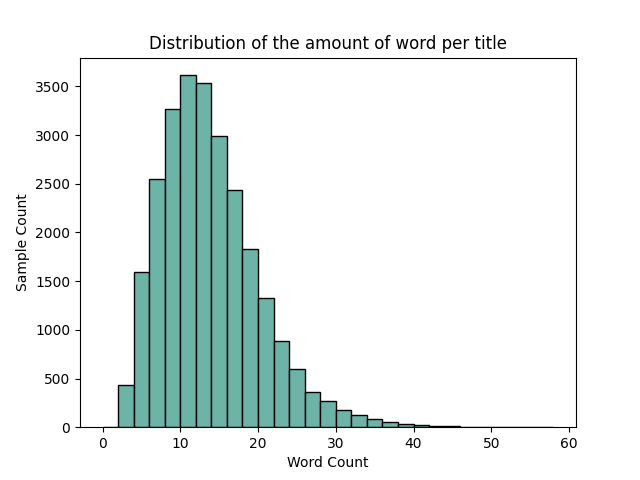}
    \caption{Distribution of the word count per title in HAL dataset, \textit{mteb\_eval} subset.}
    \label{fig:wordcount_distrib}
\end{figure}

Figure \ref{fig:extract-hal} is an extract from the HAL dataset. 
Table \ref{tab:classes-hal} lists the distribution of classes (\emph{domain} field) for the HAL dataset on \emph{raw} subset and \emph{mteb\_eval} subset, which is used for MTEB evaluation. Labels descriptions can be found at this URL: \href{https://api.archives-ouvertes.fr/ref/domain/?q=*:*&rows=393}{https://api.archives-ouvertes.fr/ref/domain/?q=*:*\&rows=393} or in Table \ref{tab:classes-hal}. After pre-processing, \emph{mteb\_eval} covers titles from 10 domains as classes with less than 500 samples were removed.
In the MTEB evaluation subset of the dataset, titles composed of 2 words or less have been removed (371 samples), resulting in an average word count of $13.4$. 
Figure \ref{fig:wordcount_distrib} shows the word count distribution per title. 
Furthermore, the dataset has been cleaned up by manually removing all non-French titles.
Additionally, it can be observed in Table \ref{tab:classes-hal} that in the original \emph{raw} dataset, the \emph{shs} and \emph{sdv} classes represent by far the majority of the dataset samples with respectively 58706 samples (73\%) and 11049 samples (13\%). 
In order to mitigate the class imbalance while preserving the majority of those classes, they have been randomly subsampled to 6701 and 4803 samples.
Furthermore, baseline models have been trained and tested to assess the usability of this dataset in other tasks, such as classification and topic modeling. 
Table \ref{tab:hal_baselines} shows the results obtained.

\begin{figure}[htbp]
    \centering
    \begin{tabular}{|c|c|p{0.4\linewidth}|}
        \hline
         \textbf{hal\_id} & \textbf{Domain} & \textbf{Title} \\ 
         \hline
         hal-02899209 & shs & 	La transformation digitale du management des ressources humaines et de ses enjeux pour les entreprises \\  
         \hline
         tel-03993881 & math & Sur l'approximation numérique de quelques problèmes en mécanique des fluides \\
        \hline
    \end{tabular}
    \caption{Extracts of HAL dataset.}
    \label{fig:extract-hal}
\end{figure}

\begin{table}[htbp]
    \centering
    \resizebox{\columnwidth}{!}{
    \begin{tabular}     
    {cccp{0.7\linewidth}}
        \hline
         \textbf{Label} & \textbf{\# \emph{raw}} & \textbf{\# \emph{mteb\_eval}} & \textbf{Description} \\ 
         \hline
         shs & 58706 & 6701 & Human and social sciences (\textit{Sciences humaines et sociales}) \\  
         sdv & 11049 & 4803 & Life science [Biology] (\textit{Sciences du vivant [Biologie]}) \\
         spi & 3601 & 3451 & Engineering science (\textit{Sciences de l'ingénieur [Physics]}) \\
         info & 3446 & 3263 & Computer Science (\textit{Informatique}) \\
         sde & 2830 & 2754 & Environment science (\textit{Sciences de l'environnement})\\
         phys & 2003 & 1926 & Physics (\textit{Physique}) \\
         sdu & 1177 & 1158 & Planet and Universe  [Physics] (\textit{Planète et Univers [Physique]})\\
         math & 862 & 824 &  Mathematics (\textit{Mathématiques}) \\
         chim & 764 & 734 & Chemistry (\textit{Chimie}) \\
         scco & 652 & 619 & Cognitive sciences (\textit{Sciences cognitives}) \\
         qfin & 183 & N/A & Economy and quantitative finance (\textit{Économie et finance quantitative} \\
         stat & 52 & N/A & Statistics (\textit{Statistiques}) \\
         other & 18 & N/A & Other (\textit{Autre}) \\
         stic & 14 & N/A & N/A\\
         nlin & 12 & N/A & Non-linear Science [Physics] (\textit{Science non linéaire [Physique]}) \\
         electromag & 3 & N/A & Electro-magnetism (\textit{Electro-magnétisme}) \\
         instrum & 2 & N/A & Instrumentation [Physics] (\textit{Instrumentation [Physique]})\\
         image & 1 & N/A & Image\\
        \hline
    \end{tabular}
    }
    \caption{Distribution of classes in HAL the \emph{raw} and \emph{mteb\_eval} subsets of the dataset.}
    \label{tab:classes-hal}
\end{table}

\begin{table}[htbp]
    \centering
    \resizebox{\columnwidth}{!}{
    \begin{tabular}{lcc}
        \hline
        \textbf{Task type} & \textbf{Model} & \textbf{Score} \\
        \multirow{3}{*}{\textbf{Classification (F1-score)}} & & \\
        \hline
        & TF-IDF + LR & $0.60 \, (\pm \, 0.002) $ \\
        & TF-IDF + SVC & $0.61 \, (\pm \, 0.001) $ \\
        & CamemBERT (fine-tuned)* & $0.6 \, (\pm \, 0.008) $ \\
        & GPT-4 (ICL)** & 0.30 \\
        \multirow{3}{*}{\textbf{Topic Modeling}} & & \\
        \hline
        & TF-IDF + LDA & 0.49 (Coherence) \\
        & & -8.23 (Perplexity)\\
        \hline
    \end{tabular}
    }
    \caption{Baselines results for HAL on a classification task and topic modeling.
    \newline
    * CamemBERT was finetuned for $5$ epochs with learning rate of $1e^{-4}$ (+ lr scheduler) and a batch size of $64$.
    \newline
    ** Due to limited budget, we evaluate \emph{GPT-4} ICL capabilities on a limited subset of our dataset (600 first samples from the test set that is generated using the same seed as for other experiments).}
    \label{tab:hal_baselines}
\end{table}

\paragraph{SummEvalFr}

Extracts of humans and machine summaries translated in French from SummEvalFr and the original ones in English from SummEval \cite{fabbri2021summeval} are shown in Figure \ref{fig:extract-summeval}.
As explained in section \ref{sec:summeval}, we use a LLM to evaluate the quality of translations for human summaries, we provide the prompt used with \emph{GPT-4} for this evaluation in Figure \ref{fig:prompt-llm-judge}.

\begin{figure}[htbp]
    \begin{lstlisting}[frame=single,linewidth=\columnwidth,breaklines=true,language=Python]
    """
    You will be given a couple of texts in English and their translation in French.
    
    Your task is to provide a 'rating' score on how well the system translated the English text into French.
    
    Give your answer as a float on a scale of 0 to 10, where 0 means that the system_translation is bad and does not represent what is being said in the original English text, and 10 means that the translation is good and represents the original English text.
    
    No need to mind the quality of the text as original English text may be of bad quality.
    
    Provide your feedback as follows:
    
    Feedback:::
    Total rating: (your rating, as a float
    between 0 and 10)
    
    Now here are the English and French texts.
    
    Original text in English: {english_text}
    Translation in French: {french_translation}
    
    Feedback:::
    Total rating:
    """
    \end{lstlisting}
    \caption{Prompt used for LLM as-judge evaluation of SummEval dataset translation.}
    \label{fig:prompt-llm-judge}
\end{figure}

Table \ref{tab:ratings} shows the distribution of ratings given by the LLM. With the scale being 10, we manually verify random samples rated above 9. We verify all samples with ratings under 9 and those with no provided rating (N/A) due to the triggering of the OpenAI content management policy. The LLM suggests that 60 samples are not correctly translated. These were verified manually, and after checking, less than 10 samples only needed to be corrected.

\begin{figure}[htbp]
    \centering
    \begin{tabular}{|p{0.2\linewidth}|p{0.3\linewidth}|p{0.3\linewidth}|}
        \hline
         \textbf{Summary type} & \textbf{Original (SummEval)} & \textbf{Translated (SummEvalFr)} \\ 
         \hline
         Human summary & \textit{The whale, Varvara, swam a round trip from Russia to Mexico, nearly 14,000 miles. The previous record was set by a humpback whale that migrated more than 10,000 miles.} & 	\textit{La baleine, Varvara, a parcouru à la nage un trajet aller-retour entre la Russie et le Mexique, soit près de 14 000 milles. Le précédent record avait été établi par une baleine à bosse qui avait migré sur plus de 10 000 miles.} \\  
         \hline
         Machine summary & \textit{north pacific gray whale has earned a spot in the record for the longest migration of a mammal ever recorded . the whale , named varvara , swam nearly 14,000 miles from the guinness worlds records . the record was set by a whale whale whale that swam a mere 10,190-mile round trip . the north coast of mexico is russian for "barbara".} & \textit{la baleine grise du pacifique nord a obtenu une place dans le record de la plus longue migration d'un mammifère jamais enregistrée. la baleine, nommée varvara, a nagé près de 14 000 miles depuis les records du monde guinness. le record a été établi par une baleine baleine qui a nagé un voyage aller-retour de seulement 10 190 miles. la côte nord du mexique est le nom russe pour "barbara".} \\
        \hline
    \end{tabular}
    \caption{Extracts of SummEvalFr dataset.}
    \label{fig:extract-summeval}
\end{figure}

\begin{table}[htbp]
    \centering
    \begin{tabular}{ccc}
         \hline
         \textbf{Quality} & \textbf{Rating} & \textbf{\# samples} \\
         \hline
         \multirow{3}{*}{Good quality} & 10.0 & 186 \\
         & 9.5 & 661 \\
         & 9.0 & 193 \\
         \hline
         \multirow{10}{*}{Not good enough} & 8.5 & 16 \\
         & 8.0 &  5 \\
         & 7.5 & 7 \\
         & 7.0 & 3 \\
         & 6.0 & 3 \\
         & 5.0 & 2 \\
         & 4.0 & 1 \\
         & 3.0 & 1 \\
         & 2.0 & 3 \\
         & N/A & 19
    \end{tabular}
    \caption{Ratings provided by the LLM judge for the quality of human summaries translations of SummEvalFr from English to French.}
    \label{tab:ratings}
\end{table}

\section{Supplementary materials for correlation analysis}

This section presents various correlations computed based on the model results on the proposed benchmark.

\begin{figure*}[htbp]
    \centering
    \includegraphics[width=1\linewidth]{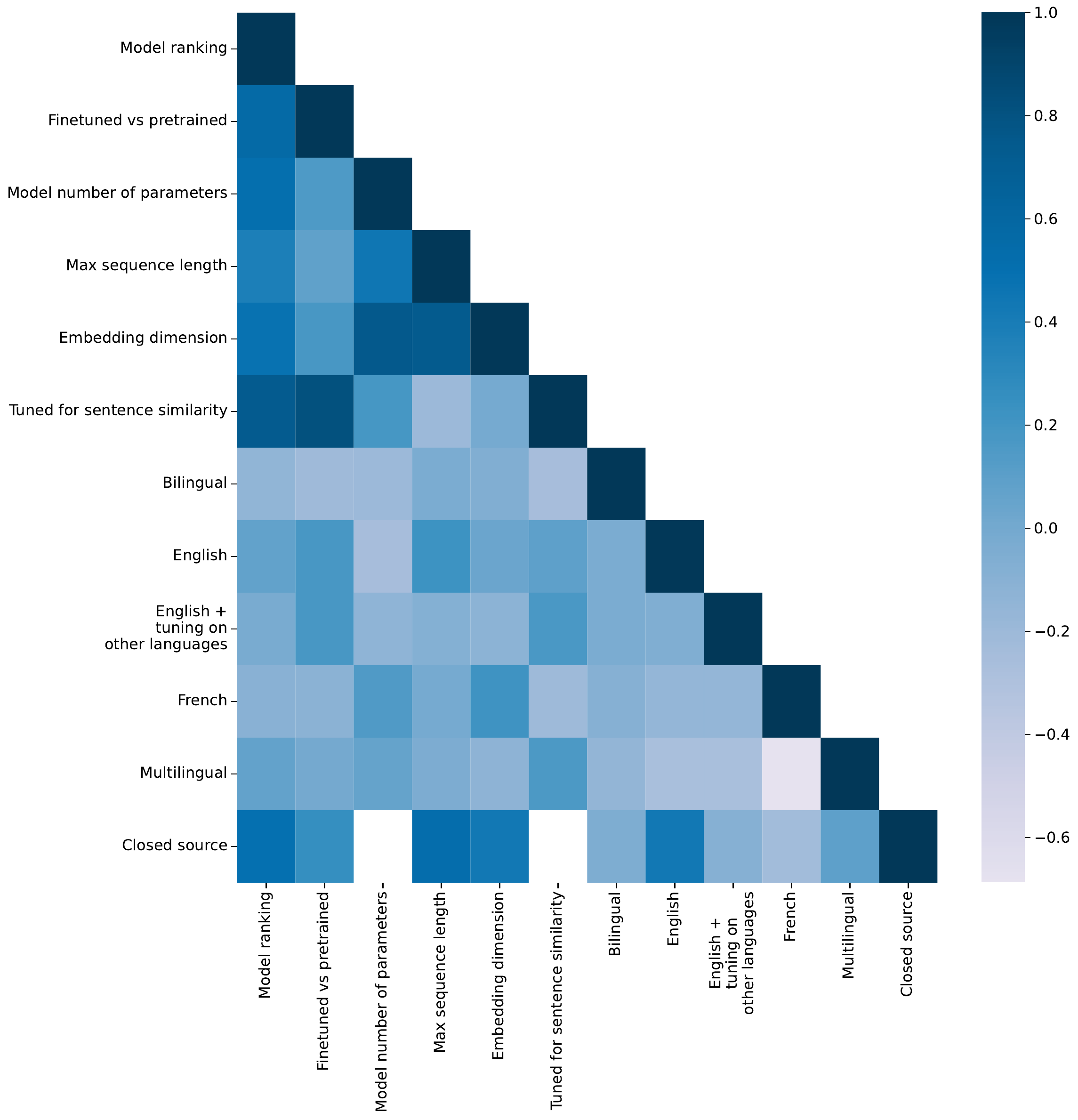}
    \caption{Heatmap representing cross-correlations between models' characteristics and models' performances.}
    \label{fig:cross_correlation_characs}
\end{figure*}

Figure \ref{fig:cross_correlation_characs} represents cross-correlations between models' performances and their studied characteristics as a heatmap.

\begin{figure*}[htbp]
    \centering
    \includegraphics[width=1\linewidth]{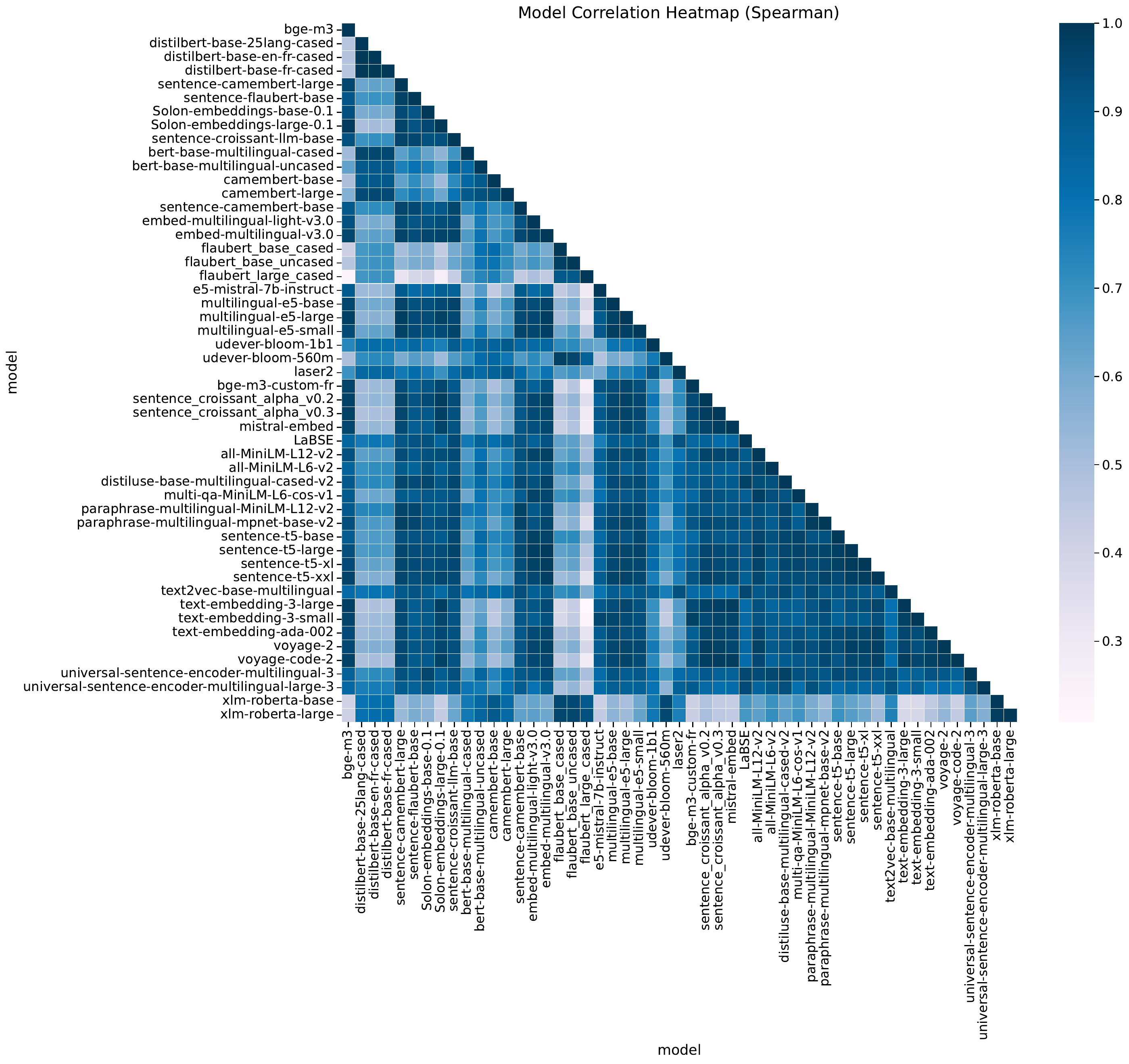}
    \caption{Heatmap representing the Spearman correlations in terms of performance across models.}
    \label{fig:correlation_model_results}
\end{figure*}

Figure \ref{fig:correlation_model_results} represents the Spearman correlations in terms of performance across models.

\begin{figure*}[htbp]
    \centering
    \includegraphics[width=1\linewidth]{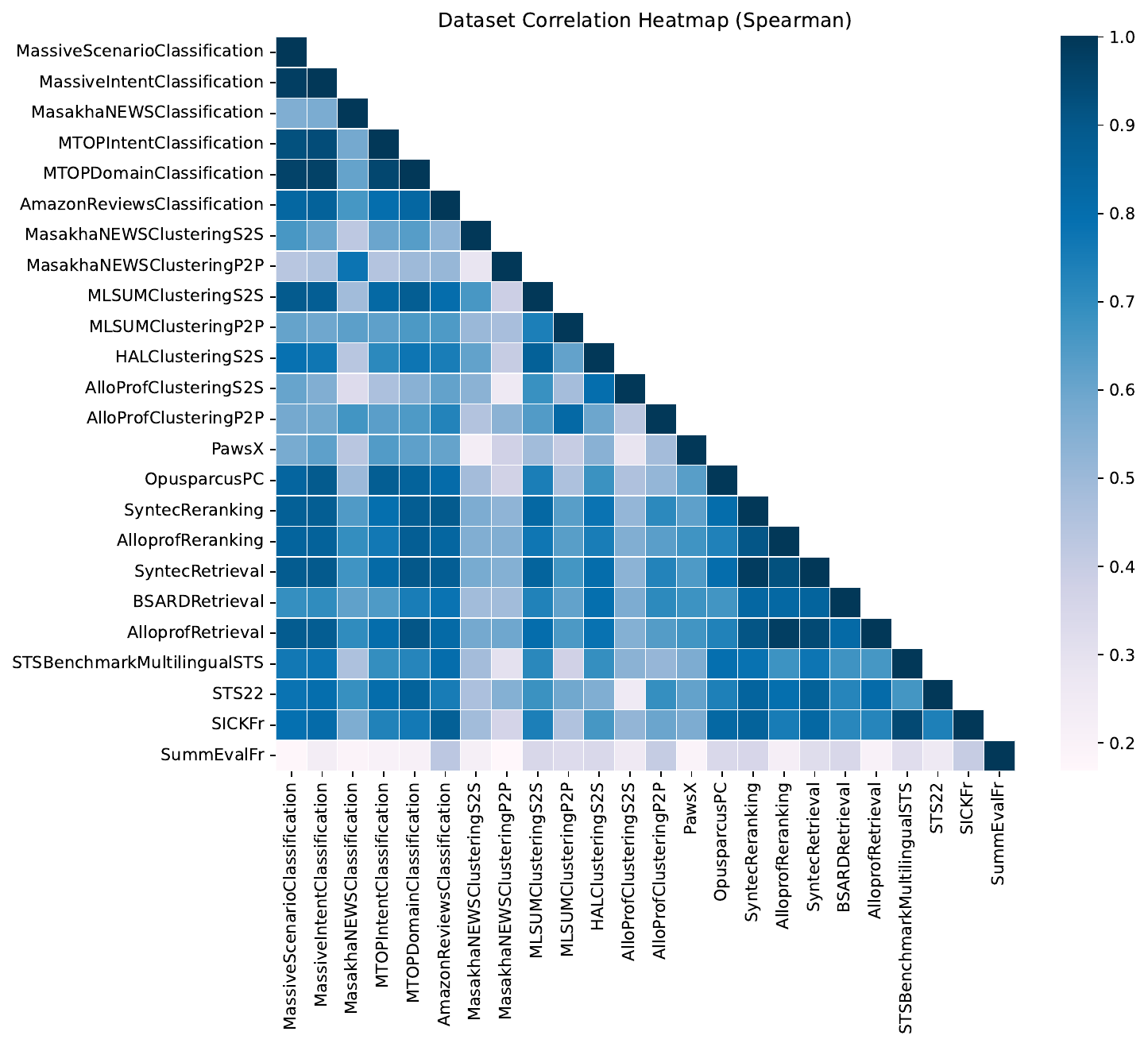}
    \caption{Heatmap representing the correlation regarding model performance across tasks.}
    \label{fig:correlation_dataset_results}
\end{figure*}

Figure \ref{fig:correlation_dataset_results} represents the Spearman correlations in terms of performance across datasets.

\section{Supplementary materials for models}
We present in this section the model characteristics we collected for the 46 evaluated models.
\input{tables/model_characteristics}

For evaluating prompt-based models such as \emph{intfloat/e5-mistral-instruct-7b}, we provide the prompts we used in Table \ref{tab:prompts}.

\begin{table*}[htbp]
    \resizebox{\textwidth}{!}{
    \begin{tabular}{c p{0.5\linewidth}}
        \hline
         \textbf{Task type} & \textbf{Prompt} \\ 
         \hline
         Classification & "Classify the following task: " \\  
         \hline
         Clustering & "Identify the topic or theme based on the text: "\\
         \hline
         Retrieval & "Retrieve semantically similar text: "\\
         \hline
         Reranking & "Re-rank the following text: "\\
         \hline
         Pair Classification & "Classify the following pair of text: "\\
         \hline
         STS & "Determine the similarity between the following text: "\\
         \hline
         Summarization & "Summarize the following text: "\\
         \hline
         Bitext Mining & "Translate the following text: "\\
        \hline
    \end{tabular}
    }
    \caption{Prompts used for the evaluation of \emph{e5-mistral-7b-instruct}.}
    \label{tab:prompts}
\end{table*}

\section{Evaluation results}

This section presents the results obtained for each model on each task. To be relevant, we used the same metrics as in MTEB, which varies from one type of task to another:
\begin{itemize}
    \item Bitext Mining: F1 score
    \item Classification: Accuracy
    \item Clustering: V measure
    \item Pair Classification: Average Precision (AP)
    \item Reranking: Mean Average Precision (MAP)
    \item Retrieval: Normalized Discounted Cumulative Gain at k (NDCG@k)
    \item STS: Spearman correlation based on cosine similarity
    \item Summarization: Spearman correlation based on cosine similarity
\end{itemize}

\subsection{Average performance per task type}
Table \ref{tab:perf_per_task_type} presents the average performance of each model on each task type.
\label{tab:appendix_avg_results}
\input{tables/average_performance_per_task}

\subsection{Evaluation results per task}

Tables \ref{tab:res_classif_pairclassif}, \ref{tab:res_reranking_retrieval} \ref{tab:res_bitext_sts_summarization} and \ref{tab:res_clustering} present the models' performance on each task type. Table \ref{tab:res_classif_pairclassif} presents the performance on classification and pair classification tasks. Table \ref{tab:res_reranking_retrieval} presents the reranking and retrieval performance. Table \ref{tab:res_bitext_sts_summarization} presents the performance on bitext mining, semantic textual similarity and summarization. Table \ref{tab:res_clustering} presents the performance on the clustering tasks.

\input{tables/results_classif_pairclassif}

\input{tables/results_reranking_retrieval}

\input{tables/results_bitext_sts_summ}

\input{tables/results_cluster}

%% file: tables/num_tokens_table.tex
\begin{table*}[htbp]
\centering
\resizebox{\textwidth}{!}{
\begin{tabular}{l|c|c|c|c}
\textbf{Dataset x Task} & \textbf{Average \# tokens} & \textbf{\# samples} & \textbf{Reference} & \textbf{License}\\
\hline
AmazonReviewsClassification & 49.6 & 5000 & \citet{amazon-review} & \emph{N/A}\\ 
MasakhaNEWSClassification & 1398.2 & 422 & \citet{Adelani2023MasakhaNEWSNT} & AFL-3.0\\ 
MassiveIntentClassification & 11.4 & 2974 & \citet{fitzgerald-etal-2023-massive} & \emph{N/A}\\ 
MassiveScenarioClassification & 11.4 & 2974 & \citet{fitzgerald-etal-2023-massive} & \emph{N/A}\\ 
MTOPDomainClassification & 12.5 & 3193 & \citet{li-etal-2021-mtop} & \emph{N/A}\\ 
MTOPIntentClassification & 12.5 & 3193 & \citet{li-etal-2021-mtop} & \emph{N/A}\\ 
AlloProfClusteringP2P & 1021.8 & 2556 & \citet{lef23} & MIT\\ 
AlloProfClusteringS2S & 8.8 & 2556 & \citet{lef23} & MIT\\ 
HALClusteringS2S & 25.6 & 26233 & \textit{Introduced by our paper} & Apache-2.0\\ 
MasakhaNEWSClusteringP2P & 1398.1 & 422 & \citet{Adelani2023MasakhaNEWSNT} & AFL-3.0\\ 
MasakhaNEWSClusteringS2S & 21.7 & 422 & \citet{Adelani2023MasakhaNEWSNT} & AFL-3.0\\ 
MLSUMClusteringP2P & 1062.1 & 15828 & \citet{scialom-etal-2020-mlsum} & Other\\ 
MLSUMClusteringS2S & 20.8 & 15828 & \citet{scialom-etal-2020-mlsum} & Other\\ 
OpusparcusPC & 9.7 & 1007 & \citet{Creutz2018Opus} & CC-BY-NC-4.0\\
PawsX & 34.9 & 2000 & \citet{yang-etal-2019-paws} & Other\\
STSBenchmarkMultilingualSTS & 18.4 & 1379 & \citet{translated_sts} & \emph{N/A}\\ 
STS22 & 722.1 & 104 & \citet{chen-etal-2022-semeval} & \emph{N/A}\\ 
SICKFr & 15.1 & 4906 & \url{https://huggingface.co/datasets/Lajavaness/SICK-fr} & Apache-2.0\\ 
DiaBLaBitextMining & 12.02 & 5748 & \citet{bawden_DiaBLa:-A-Corpus-of_2021} & CC-BY-SA-4.0\\ 
FloresBitextMining & 33.42 & 1012 & \citet{FLORES-101} & CC-BY-SA-4.0\\ 
AlloprofReranking & 48.3 - 1179.4 - 1196.4 & 2316 - 2975 - 22064 & \citet{lef23} & MIT\\
SyntecReranking & 19.2 - 402.2 - 467.2 & 100 - 100 - 917 & \textit{Introduced by our paper} & Apache-2.0\\
AlloprofRetrieval & 48.31 - 1117.91 & 2316 - 2556 & \citet{lef23} & MIT \\
BSARDRetrieval & 144.03 - 24530.8 & 222 - 22600 & \citet{louis2022statutory} & CC-BY-NC-SA-4.0 \\ 
SyntecRetrieval & 19.22 - 295.65 & 100 - 90 & \textit{Introduced by our paper} & Apache-2.0 \\ 
SummEvalFr & 657.08 - 71.18 - 107.56 & 100 - 1100 - 1600 & Created from \citet{fabbri2021summeval} & MIT \\
\end{tabular}
}
\caption{Details of the data used for each task. The average number of tokens of texts is computed using the \emph{cl100k\_base} tokenizer. For Reranking, the three numbers refer to the queries, the pairs of queries with relevant documents and the pairs of queries with irrelevant ones, respectively. The pairs of queries and documents are obtained from the 90 dataset's documents. For Retrieval datasets, the two numbers refer to the queries and the documents, respectively. For \emph{SummEvalFr}, the three numbers refer to the texts, human and machine summaries. References to all the datasets used are available.}
\label{tab:num_tokens}
\end{table*}

%% file: tables/model_characteristics.tex
\begin{table*}[htbp]
\centering
\resizebox{\textwidth}{!}{
\begin{tabular}{lcccccccc}
\textbf{Model} & \textbf{Finetuned} & \textbf{Language} & \textbf{\# params} & \textbf{Size (Gb)} & \textbf{Seq. Len.} & \textbf{Emb. dim.} & \textbf{License} & \textbf{Sentence sim}\\
\hline
bert-base-multilingual-cased & No & multilingual & 1,78e+08 & 0.71 & 512 & 768 & Apache-2.0 & No\\ 
bert-base-multilingual-uncased & No & multilingual & 1,67e+08 & 0.67 & 512 & 768 & Apache-2.0 & No\\ 
camembert-base & No & french & 1,11e+08 & 0.44 & 514 & 768 & MIT & No\\ 
camembert-large & No & french & 3,37e+08 & 1.35 & 514 & 1024 & MIT & No\\ 
sentence-camembert-base & Yes & french & 1,11e+08 & 0.44 & 128 & 768 & Apache-2.0 & Yes\\ 
sentence-camembert-large & Yes & french & 3,37e+08 & 1.35 & 514 & 1024 & Apache-2.0 & Yes\\ 
sentence-flaubert-base & Yes & french & 1,37e+08 & 0.55 & 512 & 768 & Apache-2.0 & Yes\\ 
embed-multilingual-light-v3.0 & N/A & multilingual & N/A & N/A & 512 & 384 & Closed source & N/A \\ 
embed-multilingual-v3.0 & N/A & multilingual & N/A & N/A & 512 & 1024 & Closed source & N/A \\ 
flaubert-base-cased & No & french & 1,38e+08 & 0.55 & 512 & 768 & MIT & No\\ 
flaubert-base-uncased & No & french & 1,37e+08 & 0.55 & 512 & 768 & MIT & No\\ 
flaubert-large-cased & No & french & 3,73e+08 & 1.49 & 512 & 1024 & MIT & No\\ 
distilbert-base-25lang-cased & No & multilingual & 1,08e+08 & 0.43 & 512 & 768 & Apache-2.0 & No\\ 
distilbert-base-en-fr-cased & No & bilingual & 6,86e+07 & 0.27 & 512 & 768 & Apache-2.0 & No\\ 
distilbert-base-fr-cased & No & french & 6,17e+07 & 0.25 & 512 & 768 & Apache-2.0 & No\\ 
multilingual-e5-base & No & multilingual & 2,78e+08 & 1.11 & 512 & 768 & MIT & Yes\\ 
multilingual-e5-large & No & multilingual & 5,60e+08 & 2.24 & 512 & 1024 & MIT & Yes\\ 
multilingual-e5-small & No & multilingual & 1,18e+08 & 0.47 & 512 & 384 & MIT & Yes\\ 
e5-mistral-7b-instruct & Yes & english-plus & 7,11e+09 & 28.44 & 32768 & 4096 & MIT & Yes\\ 
udever-bloom-1b1 & Yes & multilingual & 1,07e+09 & 4.26 & 2048 & 1536 & bloom-rail-1.0 & Yes\\ 
udever-bloom-560m & Yes & multilingual & 5,59e+08 & 2.24 & 2048 & 1024 & bloom-rail-1.0 & Yes\\ 
laser2 & Yes & multilingual & 4,46e+07 & 0.18 & N/A & 1024 & BSD License & Yes\\ 
all-MiniLM-L12-v2 & Yes & english-plus & 3,34e+07 & 0.13 & 128 & 384 & Apache-2.0 & Yes\\ 
all-MiniLM-L6-v2 & Yes & english-plus & 2,27e+07 & 0.09 & 256 & 384 & Apache-2.0 & Yes\\ 
distiluse-base-multilingual-cased-v2 & Yes & multilingual & 1,35e+08 & 0.54 & 128 & 512 & Apache-2.0 & Yes\\ 
LaBSE & Yes & multilingual & 4,72e+08 & 1.89 & 256 & 768 & Apache-2.0 & Yes\\ 
multi-qa-MiniLM-L6-cos-v1 & Yes & english & 2,27e+07 & 0.09 & 512 & 384 & N/A & Yes\\ 
paraphrase-multilingual-MiniLM-L12-v2 & Yes & multilingual & 1,18e+08 & 0.47 & 128 & 384 & Apache-2.0 & Yes\\ 
sentence-t5-base & Yes & multilingual & 1,10e+08 & 0.44 & 256 & 768 & Apache-2.0 & Yes\\ 
sentence-t5-large & Yes & multilingual & 3,36e+08 & 1.34 & 256 & 768 & Apache-2.0 & Yes\\ 
sentence-t5-xl & Yes & multilingual & 1,24e+09 & 4.97 & 256 & 768 & Apache-2.0 & Yes\\ 
sentence-t5-xxl & Yes & multilingual & 4,87e+09 & 19.46 & 256 & 768 & Apache-2.0 & Yes\\ 
text2vec-base-multilingual & Yes & multilingual & 1,18e+08 & 0.47 & 256 & 384 & Apache-2.0 & Yes\\ 
text-embedding-ada-002 & N/A & multilingual & N/A & N/A & 8191 & 1536 & Closed source & N/A \\ 
text-embedding-3-small & N/A & multilingual & N/A & N/A & 8191 & 1536 & Closed source & N/A \\ 
text-embedding-3-large & N/A & multilingual & N/A & N/A & 8191 & 3072 & Closed source & N/A \\ 
mistral-embed & N/A & multilingual & N/A & N/A & 16384 & 1024 & Closed source & N/A \\ 
universal-sentence-encoder-multilingual-3 & Yes & multilingual & 6,89e+07 & 0.28 & N/A & 512 & Apache-2.0 & Yes\\ 
universal-sentence-encoder-multilingual-large-3 & Yes & multilingual & 8,52e+07 & 0.34 & N/A & 512 & Apache-2.0 & Yes\\ 
xlm-roberta-base & No & multilingual & 2,78e+08 & 1.11 & 514 & 768 & MIT & No\\ 
xlm-roberta-large & No & multilingual & 5,60e+08 & 2.24 & 514 & 1024 & MIT & No\\ 
sentence-croissant-llm-base & Yes & french & 1,28e+09 & 5.12 & 256 & 2048 & MIT & Yes\\ 
paraphrase-multilingual-mpnet-base-v2 & No & multilingual & 2,78e+08 & 1.11 & 128 & 768 & Apache-2.0 & Yes\\ 
voyage-2 & N/A & english & N/A & N/A & 4000 & 1024 & Closed source & N/A \\ 
voyage-code-2 & N/A & english & N/A & N/A & 16000 & 1536 & Closed source & N/A \\
Solon-embeddings-large-0.1 & Yes & french & 5.60e+08 & 2.239561728 & 512.0 & 1024.0 & MIT & Yes \\
Solon-embeddings-base-0.1 & Yes & french & 2.78e+08 & 1.112174592 & 512.0 & 768.0 & MIT & Yes \\
sentence-croissant-alpha-v0.3 & Yes & french & 1.28e+09 & 5.11954944 & 1024.0 & 2048.0 & MIT & Yes \\
sentence-croissant-alpha-v0.2 & Yes & french & 1.28e+09 & 5.11954944 & 1024.0 & 2048.0 & MIT & Yes \\
bge-m3 & Yes & multilingual & 5.68e+08 & 2.271019008 & 8192.0 & 1024.0 & MIT & Yes \\
bge-m3-custom-fr & Yes & multilingual & 5.68e+08 & 2.271019008 & 8192.0 & 1024.0 & MIT & Yes \\
\end{tabular}
}
\caption{Models included in the benchmark with their main characteristics. The size in Gb is estimated using the number of parameters counted as float32 numbers. \emph{Sentence sim} refers to the fact that the model was trained on a task that favors semantic similarity.}
\label{tab:model_characteristics}
\end{table*}

%% file: tables/average_performance_per_task.tex
\renewcommand{\arraystretch}{1} 
\newcolumntype{R}[2]{%
    >{\adjustbox{angle=#1,lap=\width-(#2)}\bgroup}%
    l%
    <{\egroup}%
}
\newcommand*\rotate{\multicolumn{1}{R{90}{1em}}}
\begin{table*}[htbp]
\small
\centering
\resizebox{0.9\textwidth}{!}{
\begin{tabular}{lccccccccc} & \rotate{\textbf{Average}} & \rotate{\textbf{BitextMining}} & \rotate{\textbf{Classification}} & \rotate{\textbf{Clustering}} & \rotate{\textbf{PairClassification}} & \rotate{\textbf{Reranking}} & \rotate{\textbf{Retrieval}} & \rotate{\textbf{STS}} & \rotate{\textbf{Summarization}} \\ \midrule 
bge-m3 & 0.68 & 0.95 & 0.69 & 0.43 & 0.77 & 0.81 & 0.65 & 0.81 & 0.31 \\ 
distilbert-base-25lang-cased & 0.43 & 0.65 & 0.46 & 0.37 & 0.69 & 0.34 & 0.10 & 0.53 & 0.31 \\ 
distilbert-base-en-fr-cased & 0.43 & 0.65 & 0.46 & 0.38 & 0.69 & 0.34 & 0.10 & 0.54 & 0.31 \\ 
distilbert-base-fr-cased & 0.41 & 0.45 & 0.46 & 0.38 & 0.69 & 0.34 & 0.10 & 0.54 & 0.31 \\ 
sentence-camembert-large & 0.65 & 0.90 & 0.66 & 0.43 & 0.77 & 0.72 & 0.56 & 0.82 & 0.31 \\ 
sentence-flaubert-base & 0.59 & 0.80 & 0.61 & 0.41 & 0.76 & 0.65 & 0.43 & 0.79 & 0.31 \\ 
Solon-embeddings-base-0.1 & 0.64 & 0.95 & 0.67 & 0.43 & 0.76 & 0.78 & 0.41 & 0.78 & 0.31 \\ 
Solon-embeddings-large-0.1 & 0.67 & 0.96 & 0.69 & 0.42 & 0.77 & 0.79 & 0.63 & 0.80 & 0.30 \\ 
sentence-croissant-llm-base & 0.62 & 0.91 & 0.65 & 0.43 & 0.77 & 0.68 & 0.52 & 0.76 & 0.29 \\ 
bert-base-multilingual-cased & 0.44 & 0.75 & 0.46 & 0.34 & 0.70 & 0.38 & 0.10 & 0.50 & 0.29 \\ 
bert-base-multilingual-uncased & 0.49 & 0.76 & 0.48 & 0.41 & 0.70 & 0.46 & 0.19 & 0.56 & 0.31 \\ 
camembert-base & 0.35 & 0.18 & 0.42 & 0.34 & 0.68 & 0.31 & 0.02 & 0.57 & 0.30 \\ 
camembert-large & 0.37 & 0.26 & 0.49 & 0.36 & 0.65 & 0.34 & 0.07 & 0.59 & 0.17 \\ 
sentence-camembert-base & 0.57 & 0.72 & 0.57 & 0.36 & 0.74 & 0.66 & 0.43 & \textbf{0.78} & 0.29 \\ 
embed-multilingual-light-v3.0 & 0.63 & 0.89 & 0.61 & 0.39 & 0.74 & 0.76 & 0.55 & 0.78 & 0.31 \\ 
embed-multilingual-v3.0 & 0.66 & 0.94 & 0.67 & 0.41 & 0.77 & 0.79 & 0.54 & 0.81 & 0.31 \\ 
flaubert \_base \_cased & 0.34 & 0.23 & 0.25 & 0.27 & 0.67 & 0.36 & 0.08 & 0.52 & 0.31 \\ 
flaubert \_base \_uncased & 0.31 & 0.12 & 0.23 & 0.22 & 0.68 & 0.40 & 0.09 & 0.43 & 0.29 \\ 
flaubert \_large \_cased & 0.27 & 0.11 & 0.25 & 0.25 & 0.65 & 0.30 & 0.01 & 0.33 & 0.29 \\ 
e5-mistral-7b-instruct & 0.68 & 0.95 & 0.64 & 0.50 & 0.76 & 0.82 & 0.64 & 0.79 & 0.31 \\ 
multilingual-e5-base & 0.65 & 0.95 & 0.65 & 0.43 & 0.75 & 0.75 & 0.56 & 0.78 & 0.31 \\ 
multilingual-e5-large & 0.66 & 0.95 & 0.66 & 0.40 & 0.76 & 0.76 & 0.59 & 0.81 & 0.31 \\ 
multilingual-e5-small & 0.63 & 0.94 & 0.60 & 0.39 & 0.75 & 0.73 & 0.52 & 0.78 & \textbf{0.32} \\ 
udever-bloom-1b1 & 0.47 & 0.52 & 0.55 & 0.35 & 0.74 & 0.43 & 0.28 & 0.62 & 0.29 \\ 
udever-bloom-560m & 0.36 & 0.32 & 0.30 & 0.29 & 0.71 & 0.39 & 0.11 & 0.51 & 0.24 \\ 
laser2 & 0.52 & 0.95 & 0.58 & 0.30 & \textbf{0.82} & 0.44 & 0.13 & 0.67 & 0.31 \\ 
bge-m3-custom-fr & 0.66 & 0.94 & 0.67 & 0.40 & 0.77 & 0.79 & 0.59 & 0.80 & 0.30 \\ 
sentence \_croissant \_alpha \_v0.2 & 0.66 & 0.92 & 0.66 & 0.44 & 0.80 & 0.77 & 0.61 & 0.74 & 0.30 \\ 
sentence \_croissant \_alpha \_v0.3 & 0.67 & 0.92 & 0.66 & 0.46 & 0.79 & 0.78 & 0.65 & 0.77 & 0.31 \\ 
mistral-embed & 0.68 & 0.92 & 0.69 & 0.46 & 0.78 & 0.80 & \textbf{0.68} & 0.80 & 0.31 \\ 
LaBSE & 0.59 & \textbf{0.96} & 0.65 & 0.39 & 0.74 & 0.61 & 0.33 & 0.74 & 0.30 \\ 
all-MiniLM-L12-v2 & 0.51 & 0.48 & 0.52 & 0.34 & 0.72 & 0.68 & 0.43 & 0.67 & 0.27 \\ 
all-MiniLM-L6-v2 & 0.50 & 0.40 & 0.52 & 0.35 & 0.71 & 0.65 & 0.38 & 0.68 & 0.28 \\ 
distiluse-base-multilingual-cased-v2 & 0.60 & 0.94 & 0.64 & 0.39 & 0.72 & 0.69 & 0.40 & 0.75 & 0.28 \\ 
multi-qa-MiniLM-L6-cos-v1 & 0.49 & 0.38 & 0.51 & 0.33 & 0.72 & 0.64 & 0.39 & 0.67 & 0.28 \\ 
paraphrase-multilingual-MiniLM-L12-v2 & 0.60 & 0.93 & 0.60 & 0.39 & 0.74 & 0.68 & 0.44 & 0.75 & 0.29 \\ 
paraphrase-multilingual-mpnet-base-v2 & 0.63 & 0.94 & 0.63 & 0.40 & 0.76 & 0.74 & 0.50 & 0.78 & 0.30 \\ 
sentence-t5-base & 0.59 & 0.83 & 0.58 & 0.41 & 0.72 & 0.70 & 0.45 & 0.75 & 0.30 \\ 
sentence-t5-large & 0.62 & 0.90 & 0.62 & 0.42 & 0.76 & 0.73 & 0.51 & 0.75 & 0.30 \\ 
sentence-t5-xl & 0.65 & 0.91 & 0.65 & 0.43 & 0.78 & 0.76 & 0.55 & 0.77 & 0.32 \\ 
sentence-t5-xxl & 0.67 & 0.94 & 0.67 & 0.44 & 0.79 & 0.78 & 0.60 & 0.78 & 0.30 \\ 
text2vec-base-multilingual & 0.57 & 0.92 & 0.56 & 0.34 & 0.79 & 0.59 & 0.32 & 0.78 & 0.29 \\ 
text-embedding-3-large & \textbf{0.71} & \textbf{0.96} & \textbf{0.74} & 0.48 & 0.80 & \textbf{0.86} & 0.73 & 0.81 & 0.30 \\ 
text-embedding-3-small & 0.69 & 0.95 & 0.70 & 0.49 & 0.77 & 0.81 & \textbf{0.68} & 0.79 & 0.30 \\ 
text-embedding-ada-002 & 0.69 & 0.95 & 0.69 & \textbf{0.51} & 0.77 & 0.82 & 0.67 & 0.78 & 0.30 \\ 
voyage-code-2 & 0.67 & 0.86 & 0.67 & 0.47 & 0.77 & 0.81 & \textbf{0.68} & 0.78 & 0.28 \\ 
universal-sentence-encoder-multilingual-3 & 0.60 & 0.94 & 0.64 & 0.43 & 0.72 & 0.68 & 0.35 & 0.75 & 0.28 \\ 
universal-sentence-encoder-multilingual-large-3 & 0.59 & 0.95 & 0.66 & 0.37 & 0.74 & 0.67 & 0.33 & 0.74 & 0.28 \\ 
xlm-roberta-base & 0.36 & 0.48 & 0.31 & 0.28 & 0.68 & 0.30 & 0.01 & 0.51 & 0.29 \\ 
xlm-roberta-large & 0.35 & 0.35 & 0.31 & 0.29 & 0.69 & 0.35 & 0.03 & 0.49 & 0.29 
\end{tabular}
}
\caption{Average performance of models per task category.}
\label{tab:perf_per_task_type}
\end{table*}
\renewcommand{\arraystretch}{1} 

%% file: tables/results_classif_pairclassif.tex
\renewcommand{\arraystretch}{1} 
\begin{table*}[htbp]
\tiny
\centering
\resizebox{0.9\textwidth}{!}{
\begin{tabular}{l|cccccc|cc} & \rotate{\textbf{MassiveScenario}} & \rotate{\textbf{MassiveIntent}} & \rotate{\textbf{MasakhaNEWS}} & \rotate{\textbf{MTOPIntent}} & \rotate{\textbf{MTOPDomain}} & \rotate{\textbf{AmazonReviews}} & \rotate{\textbf{PawsX}} & \rotate{\textbf{OpusparcusPC}}\\ & \multicolumn{6}{c}{Classification} & \multicolumn{2}{c}{PairClassification} \\\midrule 
bge-m3 & 0.73 & 0.67 & 0.77 & 0.62 & 0.89 & 0.45 & 0.60 & 0.93 \\
distilbert-base-25lang-cased & 0.44 & 0.35 & 0.68 & 0.35 & 0.62 & 0.29 & 0.51 & 0.86 \\ 
distilbert-base-en-fr-cased & 0.44 & 0.35 & 0.68 & 0.35 & 0.62 & 0.29 & 0.51 & 0.86 \\ 
distilbert-base-fr-cased & 0.44 & 0.35 & 0.68 & 0.35 & 0.62 & 0.29 & 0.51 & 0.86 \\ 
sentence-camembert-large & 0.70 & 0.64 & 0.74 & 0.61 & 0.87 & 0.38 & 0.61 & 0.94 \\ 
sentence-flaubert-base & 0.63 & 0.59 & 0.71 & 0.53 & 0.79 & 0.40 & 0.58 & 0.93 \\ 
Solon-embeddings-base-0.1 & 0.70 & 0.65 & 0.75 & 0.62 & 0.87 & 0.41 & 0.59 & 0.93 \\ 
Solon-embeddings-large-0.1 & 0.71 & 0.67 & 0.76 & 0.69 & 0.89 & 0.42 & 0.60 & 0.94 \\ 
sentence-croissant-llm-base & 0.65 & 0.59 & 0.79 & 0.63 & 0.86 & 0.35 & 0.63 & 0.91 \\ 
bert-base-multilingual-cased & 0.44 & 0.37 & 0.64 & 0.38 & 0.64 & 0.29 & 0.53 & 0.87 \\ 
bert-base-multilingual-uncased & 0.44 & 0.38 & 0.76 & 0.39 & 0.64 & 0.29 & 0.53 & 0.87 \\ 
camembert-base & 0.39 & 0.31 & 0.66 & 0.29 & 0.58 & 0.30 & 0.52 & 0.83 \\ 
sentence-camembert-base & 0.61 & 0.52 & 0.70 & 0.43 & 0.77 & 0.36 & 0.57 & 0.92 \\ 
sentence-camembert-large & 0.69 & 0.63 & 0.81 & 0.59 & 0.86 & 0.38 & 0.60 & 0.95 \\ 
embed-multilingual-light-v3.0 & 0.59 & 0.56 & 0.83 & 0.50 & 0.81 & 0.39 & 0.57 & 0.91 \\ 
embed-multilingual-v3.0 & 0.67 & 0.63 & 0.83 & 0.61 & 0.86 & 0.42 & 0.61 & 0.94 \\ 
flaubert\_base\_cased & 0.11 & 0.07 & 0.71 & 0.09 & 0.26 & 0.25 & 0.52 & 0.82 \\ 
flaubert\_base\_uncased & 0.11 & 0.06 & 0.63 & 0.09 & 0.28 & 0.24 & 0.53 & 0.82 \\ 
flaubert\_large\_cased & 0.23 & 0.16 & 0.56 & 0.10 & 0.24 & 0.22 & 0.54 & 0.75 \\ 
e5-mistral-7b-instruct & 0.70 & 0.60 & 0.75 & 0.53 & 0.82 & 0.44 & 0.60 & 0.92 \\ 
multilingual-e5-base & 0.66 & 0.61 & 0.80 & 0.56 & 0.85 & 0.41 & 0.57 & 0.93 \\ 
multilingual-e5-large & 0.68 & 0.64 & 0.79 & 0.59 & 0.86 & 0.42 & 0.59 & 0.94 \\ 
multilingual-e5-small & 0.61 & 0.56 & 0.78 & 0.46 & 0.81 & 0.40 & 0.56 & 0.93 \\ 
udever-bloom-1b1 & 0.50 & 0.43 & 0.81 & 0.51 & 0.69 & 0.35 & 0.62 & 0.86 \\ 
udever-bloom-560m & 0.22 & 0.15 & 0.68 & 0.16 & 0.35 & 0.27 & 0.60 & 0.82 \\ laser2 & 0.59 & 0.53 & 0.66 & 0.57 & 0.76 & 0.34 & 0.70 & 0.94 \\ 
bge-m3-custom-fr & 0.75 & 0.67 & 0.70 & 0.61 & 0.90 & 0.42 & 0.61 & 0.93 \\ 
sentence\_croissant\_alpha\_v0.2 & 0.70 & 0.64 & 0.76 & 0.61 & 0.89 & 0.38 & 0.67 & 0.93 \\ 
sentence\_croissant\_alpha\_v0.3 & 0.70 & 0.65 & 0.76 & 0.59 & 0.88 & 0.36 & 0.65 & 0.93 \\ 
mistral-embed & 0.70 & 0.63 & 0.81 & 0.66 & 0.90 & 0.42 & 0.62 & 0.93 \\ 
LaBSE & 0.65 & 0.60 & 0.77 & 0.62 & 0.84 & 0.39 & 0.55 & 0.94 \\ 
all-MiniLM-L12-v2 & 0.54 & 0.45 & 0.72 & 0.39 & 0.76 & 0.28 & 0.56 & 0.87 \\ 
all-MiniLM-L6-v2 & 0.51 & 0.43 & 0.74 & 0.40 & 0.75 & 0.27 & 0.55 & 0.87 \\ distiluse-base-multilingual-cased-v2 & 0.67 & 0.60 & 0.77 & 0.56 & 0.85 & 0.36 & 0.51 & 0.92 \\ 
multi-qa-MiniLM-L6-cos-v1 & 0.50 & 0.43 & 0.76 & 0.37 & 0.73 & 0.27 & 0.57 & 0.88 \\ 
paraphrase-multilingual-MiniLM-L12-v2 & 0.65 & 0.58 & 0.76 & 0.48 & 0.78 & 0.37 & 0.57 & 0.92 \\ 
paraphrase-multilingual-mpnet-base-v2 & 0.68 & 0.62 & 0.78 & 0.52 & 0.80 & 0.40 & 0.58 & 0.93 \\ 
sentence-t5-base & 0.60 & 0.51 & 0.81 & 0.44 & 0.75 & 0.37 & 0.55 & 0.89 \\ 
sentence-t5-large & 0.64 & 0.57 & 0.80 & 0.48 & 0.80 & 0.41 & 0.60 & 0.91 \\ 
sentence-t5-xl & 0.66 & 0.61 & 0.80 & 0.54 & 0.85 & 0.44 & 0.63 & 0.92 \\ 
sentence-t5-xxl & 0.69 & 0.66 & 0.79 & 0.58 & 0.86 & 0.46 & 0.64 & 0.94 \\ 
text2vec-base-multilingual & 0.58 & 0.52 & 0.74 & 0.45 & 0.72 & 0.34 & 0.66 & 0.92 \\ 
text-embedding-3-large & 0.76 & 0.71 & 0.82 & 0.74 & 0.93 & 0.46 & 0.65 & 0.96 \\ 
text-embedding-3-small & 0.73 & 0.68 & 0.76 & 0.68 & 0.91 & 0.43 & 0.61 & 0.94 \\ 
text-embedding-ada-002 & 0.71 & 0.65 & 0.82 & 0.64 & 0.89 & 0.44 & 0.60 & 0.94 \\ 
voyage-code-2 & 0.70 & 0.63 & 0.82 & 0.59 & 0.88 & 0.42 & 0.61 & 0.93 \\ 
universal-sentence-encoder-multilingual-3 & 0.70 & 0.61 & 0.82 & 0.54 & 0.85 & 0.34 & 0.52 & 0.91 \\ 
universal-sentence-encoder-multilingual-large-3 & 0.73 & 0.66 & 0.72 & 0.64 & 0.88 & 0.35 & 0.54 & 0.93 \\ 
xlm-roberta-base & 0.23 & 0.14 & 0.60 & 0.19 & 0.44 & 0.27 & 0.51 & 0.85 \\ 
xlm-roberta-large & 0.24 & 0.16 & 0.66 & 0.15 & 0.37 & 0.27 & 0.53 & 0.84 \\ \end{tabular}
}
\caption{Performance of each model for Classification and Pair Classification.}
\label{tab:res_classif_pairclassif}
\end{table*}
\renewcommand{\arraystretch}{1} 

%% file: tables/results_reranking_retrieval.tex
\renewcommand{\arraystretch}{1}
\begin{table*}[htbp]
\tiny
\centering
\resizebox{0.9\textwidth}{!}{
\begin{tabular}{l|cc|ccc} & \rotate{\textbf{SyntecReranking}} & \rotate{\textbf{AlloprofReranking}} & \rotate{\textbf{SyntecRetrieval}} & \rotate{\textbf{BSARDRetrieval}} & \rotate{\textbf{AlloprofRetrieval}} \\ & \multicolumn{2}{c}{Reranking} & \multicolumn{3}{c}{Retrieval} \\ \midrule 
bge-m3 & 0.88 & 0.74 & 0.85 & 0.60 & 0.49 \\ 
distilbert-base-25lang-cased & 0.39 & 0.29 & 0.18 & 0.11 & 0.01 \\ 
distilbert-base-en-fr-cased & 0.39 & 0.29 & 0.18 & 0.11 & 0.01 \\ 
distilbert-base-fr-cased & 0.39 & 0.29 & 0.18 & 0.11 & 0.01 \\ 
sentence-camembert-large & 0.82 & 0.63 & 0.79 & 0.56 & 0.33 \\ 
sentence-flaubert-base & 0.81 & 0.48 & 0.69 & 0.42 & 0.18 \\ 
Solon-embeddings-base-0.1 & 0.85 & 0.71 & 0.81 & 0.00 & 0.41 \\ 
Solon-embeddings-large-0.1 & 0.87 & 0.72 & 0.85 & 0.58 & 0.47 \\ 
sentence-croissant-llm-base & 0.78 & 0.57 & 0.74 & 0.52 & 0.30 \\ 
bert-base-multilingual-cased & 0.43 & 0.32 & 0.19 & 0.10 & 0.02 \\ 
bert-base-multilingual-uncased & 0.59 & 0.33 & 0.35 & 0.16 & 0.06 \\ 
camembert-base & 0.36 & 0.26 & 0.06 & 0.00 & 0.00 \\ 
camembert-large & 0.36 & 0.33 & 0.18 & 0.01 & 0.02 \\ 
sentence-camembert-base & 0.74 & 0.58 & 0.69 & 0.39 & 0.22 \\ 
embed-multilingual-light-v3.0 & 0.82 & 0.70 & 0.77 & 0.52 & 0.35 \\ 
embed-multilingual-v3.0 & 0.84 & 0.74 & 0.79 & 0.44 & 0.38 \\ 
flaubert\_base\_cased & 0.43 & 0.29 & 0.21 & 0.02 & 0.02 \\ 
flaubert\_base\_uncased & 0.49 & 0.30 & 0.22 & 0.03 & 0.02 \\ 
flaubert\_large\_cased & 0.32 & 0.29 & 0.02 & 0.00 & 0.01 \\ 
e5-mistral-7b-instruct & 0.90 & 0.74 & 0.83 & 0.64 & 0.45 \\ 
multilingual-e5-base & 0.83 & 0.67 & 0.80 & 0.53 & 0.36 \\ 
multilingual-e5-large & 0.83 & 0.69 & 0.81 & 0.59 & 0.38 \\ 
multilingual-e5-small & 0.82 & 0.65 & 0.76 & 0.52 & 0.27 \\ 
udever-bloom-1b1 & 0.48 & 0.39 & 0.41 & 0.32 & 0.12 \\ 
udever-bloom-560m & 0.47 & 0.31 & 0.24 & 0.06 & 0.02 \\ 
laser2 & 0.49 & 0.39 & 0.29 & 0.08 & 0.03 \\ 
bge-m3-custom-fr & 0.85 & 0.74 & 0.79 & 0.53 & 0.45 \\ 
sentence\_croissant\_alpha\_v0.2 & 0.82 & 0.72 & 0.79 & 0.60 & 0.45 \\ 
sentence\_croissant\_alpha\_v0.3 & 0.82 & 0.74 & 0.80 & 0.66 & 0.49 \\ 
mistral-embed & 0.81 & 0.78 & 0.79 & 0.68 & 0.57 \\ 
LaBSE & 0.68 & 0.55 & 0.55 & 0.23 & 0.20 \\ 
all-MiniLM-L12-v2 & 0.69 & 0.67 & 0.61 & 0.34 & 0.33 \\ 
all-MiniLM-L6-v2 & 0.67 & 0.63 & 0.60 & 0.27 & 0.28 \\ 
distiluse-base-multilingual-cased-v2 & 0.75 & 0.62 & 0.65 & 0.29 & 0.27 \\ 
multi-qa-MiniLM-L6-cos-v1 & 0.65 & 0.63 & 0.58 & 0.30 & 0.30 \\ 
paraphrase-multilingual-MiniLM-L12-v2 & 0.73 & 0.62 & 0.66 & 0.38 & 0.27 \\ 
paraphrase-multilingual-mpnet-base-v2 & 0.81 & 0.67 & 0.76 & 0.43 & 0.31 \\ 
sentence-t5-base & 0.76 & 0.63 & 0.67 & 0.40 & 0.28 \\ 
sentence-t5-large & 0.78 & 0.68 & 0.71 & 0.47 & 0.35 \\ 
sentence-t5-xl & 0.81 & 0.71 & 0.74 & 0.50 & 0.40 \\ 
sentence-t5-xxl & 0.82 & 0.75 & 0.79 & 0.56 & 0.46 \\ 
text2vec-base-multilingual & 0.63 & 0.56 & 0.50 & 0.26 & 0.19 \\ 
text-embedding-3-large & 0.92 & 0.80 & 0.87 & 0.73 & 0.60 \\ 
text-embedding-3-small & 0.89 & 0.74 & 0.87 & 0.66 & 0.52 \\ 
text-embedding-ada-002 & 0.89 & 0.76 & 0.86 & 0.64 & 0.52 \\ 
voyage-code-2 & 0.87 & 0.76 & 0.83 & 0.68 & 0.53 \\ 
universal-sentence-encoder-multilingual-3 & 0.74 & 0.62 & 0.70 & 0.00 & 0.35 \\ 
universal-sentence-encoder-multilingual-large-3 & 0.69 & 0.64 & 0.64 & 0.00 & 0.34 \\ 
xlm-roberta-base & 0.32 & 0.28 & 0.03 & 0.00 & 0.00 \\ 
xlm-roberta-large & 0.39 & 0.31 & 0.07 & 0.01 & 0.01 \\ \end{tabular}
}
\caption{Performance of each model for Retrieval and Reranking.}
\label{tab:res_reranking_retrieval}
\end{table*}
\renewcommand{\arraystretch}{1} 

%% file: tables/results_bitext_sts_summ.tex
\renewcommand{\arraystretch}{1} 
\begin{table*}[htbp]
\tiny
\centering
\resizebox{0.9\textwidth}{!}{
\begin{tabular}{l|ccc|ccc|c} & \rotate{\textbf{Flores\_fr-en}} & \rotate{\textbf{Flores\_en-fr}} & \rotate{\textbf{DiaBla\_fr-en}} & \rotate{\textbf{STSBenchmarkMultilingual}} & \rotate{\textbf{STS22}} & \rotate{\textbf{SICKFr}} & \rotate{\textbf{SummEvalFr}} \\ & \multicolumn{3}{c}{BitextMining} & \multicolumn{3}{c}{STS} & Summarization \\ \midrule 
bge-m3 & 1.00 & 1.00 & 0.85 & 0.82 & 0.82 & 0.78 & 0.31 \\ 
distilbert-base-25lang-cased & 0.92 & 0.91 & 0.11 & 0.57 & 0.41 & 0.62 & 0.31 \\ 
distilbert-base-en-fr-cased & 0.92 & 0.91 & 0.11 & 0.57 & 0.42 & 0.62 & 0.31 \\ 
distilbert-base-fr-cased & 0.63 & 0.65 & 0.06 & 0.57 & 0.43 & 0.62 & 0.31 \\ 
sentence-camembert-large & 0.99 & 1.00 & 0.70 & 0.86 & 0.82 & 0.78 & 0.31 \\ 
sentence-flaubert-base & 0.96 & 0.97 & 0.47 & 0.86 & 0.74 & 0.78 & 0.31 \\ 
Solon-embeddings-base-0.1 & 1.00 & 1.00 & 0.85 & 0.79 & 0.81 & 0.75 & 0.31 \\ 
Solon-embeddings-large-0.1 & 1.00 & 1.00 & 0.87 & 0.80 & 0.83 & 0.77 & 0.30 \\ 
sentence-croissant-llm-base & 1.00 & 1.00 & 0.74 & 0.79 & 0.79 & 0.70 & 0.29 \\ 
bert-base-multilingual-cased & 0.97 & 0.98 & 0.30 & 0.52 & 0.39 & 0.59 & 0.29 \\ 
bert-base-multilingual-uncased & 0.95 & 0.98 & 0.36 & 0.55 & 0.56 & 0.58 & 0.31 \\ 
camembert-base & 0.26 & 0.25 & 0.04 & 0.55 & 0.61 & 0.54 & 0.30 \\ 
sentence-camembert-base & 0.90 & 0.90 & 0.36 & 0.82 & 0.78 & 0.74 & 0.29 \\ 
sentence-camembert-large & 0.99 & 1.00 & 0.68 & 0.86 & 0.82 & 0.78 & 0.31 \\ 
embed-multilingual-light-v3.0 & 1.00 & 1.00 & 0.66 & 0.76 & 0.83 & 0.76 & 0.31 \\ 
embed-multilingual-v3.0 & 1.00 & 1.00 & 0.83 & 0.82 & 0.83 & 0.79 & 0.31 \\ 
flaubert\_base\_cased & 0.31 & 0.36 & 0.02 & 0.37 & 0.65 & 0.54 & 0.31 \\ 
flaubert\_base\_uncased & 0.25 & 0.08 & 0.03 & 0.33 & 0.55 & 0.42 & 0.29 \\
flaubert\_large\_cased & 0.15 & 0.17 & 0.01 & 0.16 & 0.49 & 0.35 & 0.29 \\ 
e5-mistral-7b-instruct & 1.00 & 1.00 & 0.85 & 0.83 & 0.76 & 0.79 & 0.31 \\ 
multilingual-e5-base & 1.00 & 1.00 & 0.85 & 0.81 & 0.78 & 0.76 & 0.31 \\ 
multilingual-e5-large & 1.00 & 1.00 & 0.85 & 0.83 & 0.80 & 0.79 & 0.31 \\ 
multilingual-e5-small & 1.00 & 1.00 & 0.82 & 0.79 & 0.80 & 0.76 & 0.32 \\ 
udever-bloom-1b1 & 0.75 & 0.78 & 0.03 & 0.50 & 0.77 & 0.60 & 0.29 \\ 
udever-bloom-560m & 0.50 & 0.37 & 0.08 & 0.37 & 0.61 & 0.55 & 0.24 \\ 
laser2 & 1.00 & 1.00 & 0.86 & 0.70 & 0.65 & 0.65 & 0.31 \\ 
bge-m3-custom-fr & 1.00 & 1.00 & 0.83 & 0.81 & 0.82 & 0.76 & 0.30 \\ 
sentence\_croissant\_alpha\_v0.2 & 1.00 & 1.00 & 0.75 & 0.73 & 0.79 & 0.69 & 0.30 \\ 
sentence\_croissant\_alpha\_v0.3 & 1.00 & 1.00 & 0.77 & 0.78 & 0.81 & 0.72 & 0.31 \\ 
mistral-embed & 1.00 & 1.00 & 0.75 & 0.80 & 0.83 & 0.76 & 0.31 \\ 
LaBSE & 1.00 & 1.00 & 0.88 & 0.75 & 0.78 & 0.70 & 0.30 \\ 
all-MiniLM-L12-v2 & 0.71 & 0.62 & 0.10 & 0.67 & 0.70 & 0.63 & 0.27 \\ 
all-MiniLM-L6-v2 & 0.62 & 0.56 & 0.03 & 0.65 & 0.77 & 0.62 & 0.28 \\ 
distiluse-base-multilingual-cased-v2 & 1.00 & 1.00 & 0.83 & 0.77 & 0.76 & 0.72 & 0.28 \\ 
multi-qa-MiniLM-L6-cos-v1 & 0.55 & 0.50 & 0.09 & 0.64 & 0.75 & 0.62 & 0.28 \\ 
paraphrase-multilingual-MiniLM-L12-v2 & 1.00 & 1.00 & 0.78 & 0.80 & 0.71 & 0.75 & 0.29 \\ 
paraphrase-multilingual-mpnet-base-v2 & 1.00 & 1.00 & 0.81 & 0.85 & 0.74 & 0.76 & 0.30 \\ 
sentence-t5-base & 0.97 & 0.96 & 0.55 & 0.74 & 0.78 & 0.72 & 0.30 \\ 
sentence-t5-large & 0.99 & 0.99 & 0.71 & 0.78 & 0.75 & 0.73 & 0.30 \\ 
sentence-t5-xl & 0.99 & 0.99 & 0.76 & 0.79 & 0.77 & 0.75 & 0.32 \\ 
sentence-t5-xxl & 1.00 & 1.00 & 0.83 & 0.81 & 0.77 & 0.77 & 0.30 \\ 
text2vec-base-multilingual & 0.99 & 0.99 & 0.78 & 0.83 & 0.74 & 0.77 & 0.29 \\ 
text-embedding-3-large & 1.00 & 1.00 & 0.88 & 0.83 & 0.82 & 0.79 & 0.30 \\ 
text-embedding-3-small & 1.00 & 1.00 & 0.86 & 0.81 & 0.81 & 0.76 & 0.30 \\ 
text-embedding-ada-002 & 0.99 & 0.99 & 0.86 & 0.78 & 0.81 & 0.76 & 0.30 \\ 
voyage-code-2 & 1.00 & 0.99 & 0.60 & 0.79 & 0.80 & 0.74 & 0.28 \\ 
universal-sentence-encoder-multilingual-3 & 1.00 & 1.00 & 0.82 & 0.75 & 0.78 & 0.71 & 0.28 \\ 
universal-sentence-encoder-multilingual-large-3 & 1.00 & 1.00 & 0.84 & 0.78 & 0.71 & 0.74 & 0.28 \\ 
xlm-roberta-base & 0.70 & 0.53 & 0.21 & 0.46 & 0.57 & 0.49 & 0.29 \\ 
xlm-roberta-large & 0.65 & 0.26 & 0.13 & 0.42 & 0.55 & 0.50 & 0.29 \\ 
\end{tabular}
}
\caption{Performance of each model for Bitext Mining, Semantic Textual Similarity (STS) and Summarization.}
\label{tab:res_bitext_sts_summarization}
\end{table*}
\renewcommand{\arraystretch}{1} 

%% file: tables/results_cluster.tex
\renewcommand{\arraystretch}{1} 
\begin{table*}[htbp]
\tiny 
\centering
\resizebox{0.9\textwidth}{!}{
\begin{tabular}{l|ccccccc} & \rotate{\textbf{MasakhaNEWSS2S}} & \rotate{\textbf{MasakhaNEWSP2P}} & \rotate{\textbf{MLSUMS2S}} & \rotate{\textbf{MLSUMP2P}} & \rotate{\textbf{HALS2S}} & \rotate{\textbf{AlloProfS2S}} & \rotate{\textbf{AlloProfP2P}} \\  & \multicolumn{7}{c}{Clustering} \\ \midrule
bge-m3 & 0.42 & 0.45 & 0.44 & 0.43 & 0.31 & 0.37 & 0.59 \\
distilbert-base-25lang-cased & 0.33 & 0.32 & 0.31 & 0.41 & 0.24 & 0.43 & 0.57 \\ 
distilbert-base-en-fr-cased & 0.34 & 0.34 & 0.31 & 0.41 & 0.25 & 0.42 & 0.57 \\ 
distilbert-base-fr-cased & 0.35 & 0.34 & 0.31 & 0.41 & 0.24 & 0.43 & 0.57 \\ 
sentence-camembert-large & 0.37 & 0.44 & 0.43 & 0.43 & 0.32 & 0.40 & 0.62 \\ 
sentence-flaubert-base & 0.30 & 0.49 & 0.41 & 0.41 & 0.32 & 0.40 & 0.57 \\ 
Solon-embeddings-base-0.1 & 0.36 & 0.50 & 0.42 & 0.43 & 0.30 & 0.37 & 0.61 \\ 
Solon-embeddings-large-0.1 & 0.31 & 0.46 & 0.43 & 0.43 & 0.32 & 0.37 & 0.63 \\
sentence-croissant-llm-base & 0.41 & 0.54 & 0.34 & 0.43 & 0.29 & 0.33 & 0.64 \\ 
bert-base-multilingual-cased & 0.24 & 0.24 & 0.32 & 0.41 & 0.25 & 0.43 & 0.51 \\ 
bert-base-multilingual-uncased & 0.42 & 0.50 & 0.31 & 0.43 & 0.26 & 0.35 & 0.61 \\ 
camembert-base & 0.27 & 0.44 & 0.27 & 0.41 & 0.16 & 0.29 & 0.54 \\ 
camembert-large & 0.33 & 0.42 & 0.35 & 0.44 & 0.03 & 0.34 & 0.59 \\ 
sentence-camembert-base & 0.31 & 0.36 & 0.27 & 0.36 & 0.25 & 0.39 & 0.59 \\ 
embed-multilingual-light-v3.0 & 0.29 & 0.57 & 0.33 & 0.43 & 0.20 & 0.31 & 0.62 \\ 
embed-multilingual-v3.0 & 0.32 & 0.53 & 0.35 & 0.45 & 0.24 & 0.36 & 0.64 \\ 
flaubert\_base\_cased & 0.21 & 0.42 & 0.17 & 0.39 & 0.04 & 0.14 & 0.53 \\ 
flaubert\_base\_uncased & 0.23 & 0.28 & 0.15 & 0.33 & 0.02 & 0.13 & 0.43 \\ 
flaubert\_large\_cased & 0.25 & 0.26 & 0.19 & 0.38 & 0.07 & 0.22 & 0.41 \\ 
e5-mistral-7b-instruct & 0.65 & 0.38 & 0.44 & 0.45 & 0.37 & 0.58 & 0.64 \\ 
multilingual-e5-base & 0.51 & 0.48 & 0.39 & 0.43 & 0.28 & 0.33 & 0.62 \\ 
multilingual-e5-large & 0.31 & 0.41 & 0.38 & 0.44 & 0.28 & 0.32 & 0.63 \\ 
multilingual-e5-small & 0.39 & 0.40 & 0.38 & 0.43 & 0.21 & 0.33 & 0.61 \\ 
udever-bloom-1b1 & 0.27 & 0.40 & 0.30 & 0.44 & 0.16 & 0.27 & 0.62 \\ 
udever-bloom-560m & 0.21 & 0.38 & 0.25 & 0.36 & 0.08 & 0.22 & 0.54 \\ 
laser2 & 0.30 & 0.32 & 0.27 & 0.35 & 0.12 & 0.26 & 0.48 \\ 
bge-m3-custom-fr & 0.42 & 0.29 & 0.42 & 0.42 & 0.31 & 0.39 & 0.58 \\ 
sentence\_croissant\_alpha\_v0.2 & 0.32 & 0.56 & 0.44 & 0.45 & 0.33 & 0.38 & 0.62 \\ 
sentence\_croissant\_alpha\_v0.3 & 0.38 & 0.58 & 0.44 & 0.44 & 0.35 & 0.41 & 0.60 \\ 
mistral-embed & 0.40 & 0.48 & 0.43 & 0.45 & 0.35 & 0.49 & 0.62 \\ 
LaBSE & 0.38 & 0.46 & 0.35 & 0.42 & 0.25 & 0.32 & 0.55 \\ 
all-MiniLM-L12-v2 & 0.32 & 0.43 & 0.29 & 0.34 & 0.25 & 0.32 & 0.46 \\ 
all-MiniLM-L6-v2 & 0.41 & 0.35 & 0.28 & 0.37 & 0.23 & 0.32 & 0.52 \\ 
distiluse-base-multilingual-cased-v2 & 0.33 & 0.54 & 0.35 & 0.40 & 0.22 & 0.35 & 0.56 \\ 
multi-qa-MiniLM-L6-cos-v1 & 0.27 & 0.54 & 0.26 & 0.35 & 0.14 & 0.26 & 0.49 \\ 
paraphrase-multilingual-MiniLM-L12-v2 & 0.34 & 0.37 & 0.37 & 0.40 & 0.30 & 0.42 & 0.56 \\ 
paraphrase-multilingual-mpnet-base-v2 & 0.31 & 0.42 & 0.38 & 0.41 & 0.31 & 0.45 & 0.54 \\ 
sentence-t5-base & 0.36 & 0.62 & 0.30 & 0.41 & 0.22 & 0.36 & 0.58 \\ 
sentence-t5-large & 0.31 & 0.59 & 0.32 & 0.42 & 0.25 & 0.40 & 0.62 \\ 
sentence-t5-xl & 0.32 & 0.63 & 0.34 & 0.42 & 0.27 & 0.41 & 0.60 \\ 
sentence-t5-xxl & 0.38 & 0.61 & 0.35 & 0.42 & 0.30 & 0.44 & 0.61 \\ 
text2vec-base-multilingual & 0.33 & 0.39 & 0.30 & 0.36 & 0.21 & 0.33 & 0.49 \\ 
text-embedding-3-large & 0.40 & 0.53 & 0.46 & 0.46 & 0.37 & 0.54 & 0.62 \\ 
text-embedding-3-small & 0.55 & 0.45 & 0.46 & 0.46 & 0.36 & 0.51 & 0.61 \\ 
text-embedding-ada-002 & 0.49 & 0.68 & 0.42 & 0.45 & 0.35 & 0.54 & 0.65 \\ 
voyage-code-2 & 0.35 & 0.57 & 0.41 & 0.45 & 0.35 & 0.51 & 0.62 \\ 
universal-sentence-encoder-multilingual-3 & 0.40 & 0.61 & 0.36 & 0.44 & 0.24 & 0.38 & 0.57 \\ 
universal-sentence-encoder-multilingual-large-3 & 0.40 & 0.24 & 0.38 & 0.41 & 0.23 & 0.38 & 0.54 \\ 
xlm-roberta-base & 0.24 & 0.29 & 0.24 & 0.40 & 0.09 & 0.20 & 0.52 \\ 
xlm-roberta-large & 0.22 & 0.34 & 0.19 & 0.43 & 0.06 & 0.21 & 0.57 \\ \bottomrule
\end{tabular}
}
\caption{Performance of each model for Clustering.}
\label{tab:res_clustering}
\end{table*}
\renewcommand{\arraystretch}{1} 